# On-line Planning and Scheduling:
# An Application to Controlling Modular Printers

**Wheeler Ruml**                                      RUML AT CS.UNH.EDU
*Department of Computer Science*
*University of New Hampshire*
*33 Academic Way*
*Durham, NH 03824 USA*

**Minh Binh Do**                                      MINHDO AT PARC.COM
**Rong Zhou**                                         RZHOU AT PARC.COM
**Markus P. J. Fromherz**                             FROMHERZ AT PARC.COM
*Palo Alto Research Center*
*3333 Coyote Hill Road*
*Palo Alto, CA 94304 USA*

## Abstract

We present a case study of artificial intelligence techniques applied to the control of production printing equipment. Like many other real-world applications, this complex domain requires high-speed autonomous decision-making and robust continual operation. To our knowledge, this work represents the first successful industrial application of embedded domain-independent temporal planning. Our system handles execution failures and multi-objective preferences. At its heart is an on-line algorithm that combines techniques from state-space planning and partial-order scheduling. We suggest that this general architecture may prove useful in other applications as more intelligent systems operate in continual, on-line settings. Our system has been used to drive several commercial prototypes and has enabled a new product architecture for our industrial partner. When compared with state-of-the-art off-line planners, our system is hundreds of times faster and often finds better plans. Our experience demonstrates that domain-independent AI planning based on heuristic search can flexibly handle time, resources, replanning, and multiple objectives in a high-speed practical application without requiring hand-coded control knowledge.

## 1. Introduction

It is a sustaining goal of artificial intelligence to develop techniques enabling autonomous agents to robustly achieve multiple interacting goals in a dynamic environment. This goal is not just intellectually attractive. It also happens to align perfectly with the needs of many commercial manufacturing plants. In this paper, we focus on one particular manufacturing setting: high-speed digital production printing systems. These large machines use xerography to print the requested images on individual sheets of paper. Unlike traditional continuous-feed offset presses, digital printers can treat each sheet differently: feeding different types and sizes of media, printing different kinds of images, and performing different preparatory and finishing operations. Often, a single integrated machine can transform blank sheets into a complete document, such as a bound book or a folded bill in a sealed envelope. It is sometimes even possible to process different kinds of jobs simultaneously on the same equipment. A printer controller must plan quickly and reliably; otherwise expensive human intervention will be required. Designing a high-performance yet cost-effective controller for such machines is made more difficult by the current trend towards increased modularity, in which each customer's system is unique and includes only those components that are most appropriate for their needs. We have been working closely with the Xerox Corporation to explore architectures in which printing systems can be composed of literally hundreds of modules, possibly including multiple specialized printing modules, working together at high speed.

In this paper, we demonstrate how techniques from artificial intelligence can be used to control such machines. Requests for print jobs become goals for the system to achieve, the various actuators and mechanisms in the machine become actions and resources to be used in achieving these goals, and sensors provide feedback on action execution and the state of the system. To provide high productivity (and thus high return on investment for the equipment owner), the planning and control techniques must be fast and produce optimal or near-optimal plans. To reduce the need for operator oversight and to allow the use of very complex mechanisms, the system must be as autonomous and autonomic as possible. Because operators can make mistakes and even highly-engineered system modules can fail, the system must cope with execution failure and unexpected events. And because the system must work with legacy modules in order to be commercially viable, its architecture must tolerate components that are out of its direct control.

To meet these requirements, we present a novel architecture for on-line planning, execution, and replanning that synthesizes techniques from state-space planning (Ghallab, Nau, & Traverso, 2004) and partial-order scheduling (Smith & Cheng, 1993). We develop new heuristic evaluation functions for temporal planning that incorporate some of the effects of resource constraints. Although domain-independent AI planning is often regarded as too expensive for use in a soft real-time setting, our system achieves good performance without any hand-coded control rules, despite the additional requirements of reasoning about temporal actions and resources. By avoiding domain-dependent search control knowledge, it becomes possible to use the same planner to run very different printing systems at full productivity. The success of our system has enabled a new modular product architecture that can span multiple markets. Much as previous work brought constraint-based scheduling into daily use in print shops and offices world-wide (Fromherz, Saraswat, & Bobrow, 1999; Fromherz, Bobrow, & de Kleer, 2003), our work can bring domain-independent temporal planning into continual widespread use by everyday people. Our approach is practical and efficient, and it showcases the flexibility inherent in viewing planning as heuristic search.

After discussing the application context in more detail, we will present an overview of our system, followed by detailed discussion of its major aspects: nominal planning, exception handling, and multiple objectives. As we go, we will present empirical measurements demonstrating that large printing systems can be controlled by our system while meeting our real-time requirements. In particular, Section 4.4.1 describes a comparison with state-of-the-art generic off-line planners that demonstrates that our planner finds plans hundreds of times faster that are often of higher quality, and an on-line appendix provides videos of our planner controlling our hardware prototype. Our integrated approach to on-line planning and scheduling allows us to achieve high throughput even for complex systems.

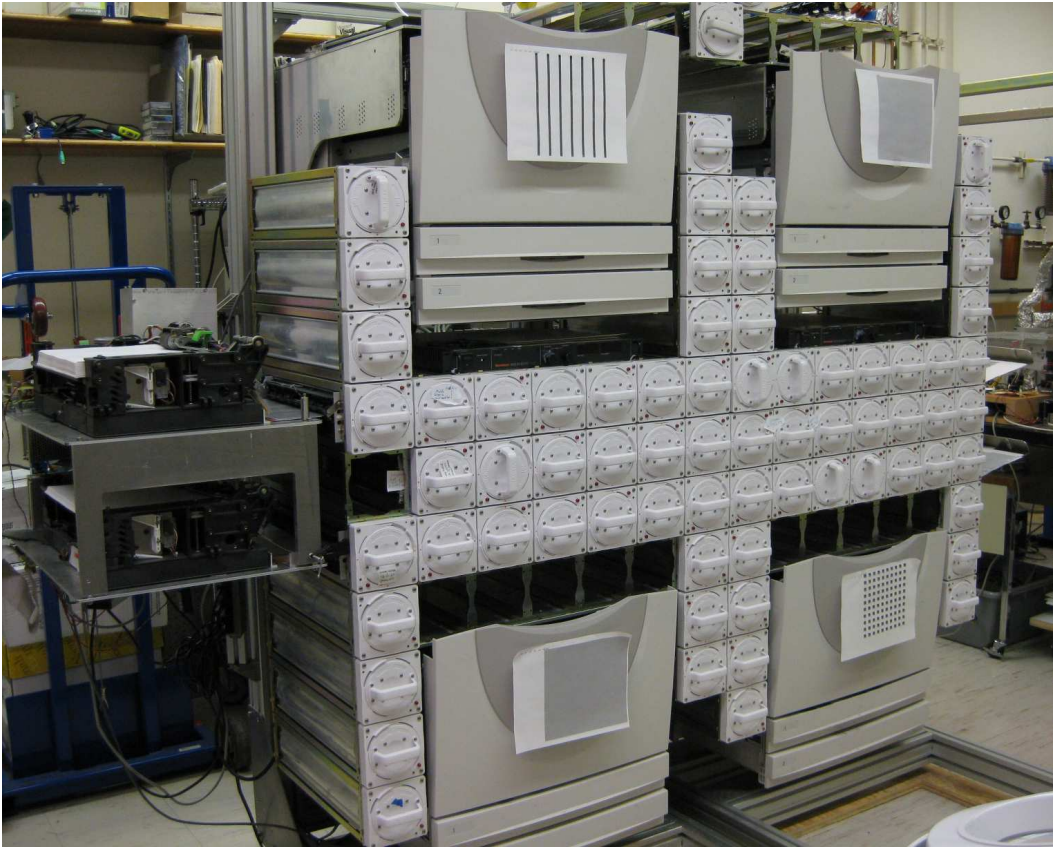

Figure 1: A prototype modular printer built at PARC. The system is composed of approximately 170 individually controlled modules, including four print engines.

We conclude the paper with a summary of general lessons we derived from building this application.

## 2. Application Context

In analogy to other parallel systems such as RAID storage, our approach to modular printing systems is called Rack Mounted Printing (RMP). An RMP system can be seen as a network of transports linking multiple printing engines. These transports are known as the media path. Figure 1 shows a four-engine prototype printer built at the Palo Alto Research Center (PARC) from over 170 independently controlled modules. Figure 2 provides a schematic side view, showing the many possible paper paths linking the paper feeders to the possible output trays. (Video 1 in the on-line appendix, 'nominal simulation,' presents an animation of Figure 2.) Multiple feeders allow blank sheets to enter the printer at a high rate and multiple finishers allow several print jobs to run simultaneously. Having redundant paths through the machine enables graceful degradation of performance when modules fail. By building the system out of relatively small modules, we enable easy reconfiguration of the components to add new modules and functionality. Achieving these benefits, however, poses a considerable control challenge.

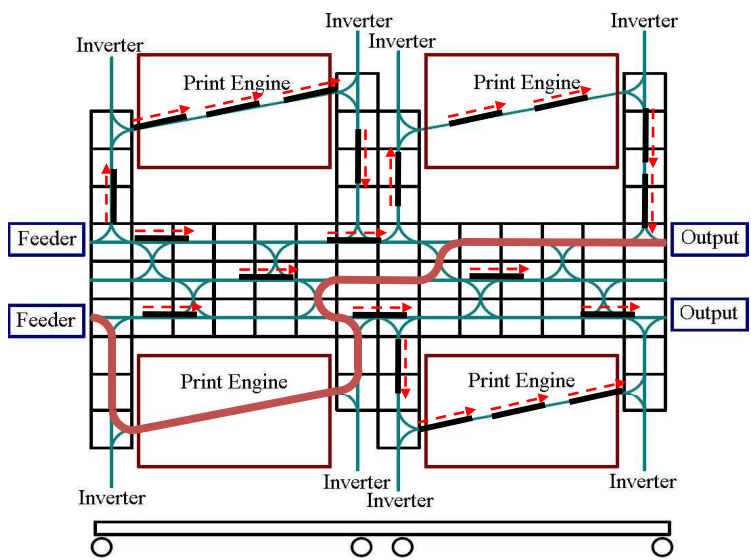

Figure 2: A schematic side view of the modular printer indicating the feeders, paper path, and output trays.

The modular printing domain is reminiscent of 'mass customization,' in which mass-produced products are closely tailored and personalized to individual customers' needs. It is also similar to package routing or logistics problems. From a control perspective, it involves planning and scheduling a series of sheet requests that arrive asynchronously over time from the front-end print-job submission and rendering engine. The system runs at high speed, with several sheet requests arriving per second, possibly for many hours. Each sheet request completely describes the attributes of a desired final product. There may be several different sequences of actions that can be used to print a given sheet. For example, in Figure 2, a blank sheet may be fed from either of the two feeders, then routed to any one of the four print engines (or through any combination of two of the four engines in the case of duplex printing) and then to either finisher (unless the sheet is part of an on-going print job).

This on-line planning problem is complicated by the fact that many sheets are in-flight simultaneously and the plan for the new sheet must not interfere with those sheets. Most actions require the use of physical printer components, so planning for later sheets must take into account the resource commitments in plans already released for production. Because modern printers are highly configurable, can execute an large variety of jobs potentially simultaneously, and have a large variety of constraints on feasible plans, hard-coded or locally-reactive plans do not suffice (Fromherz et al., 1999). In fact, printer engineers at Xerox delight in uncovering situations in which products from competing manufacturers, who do not use model-based planning, attempt to execute infeasible plans.

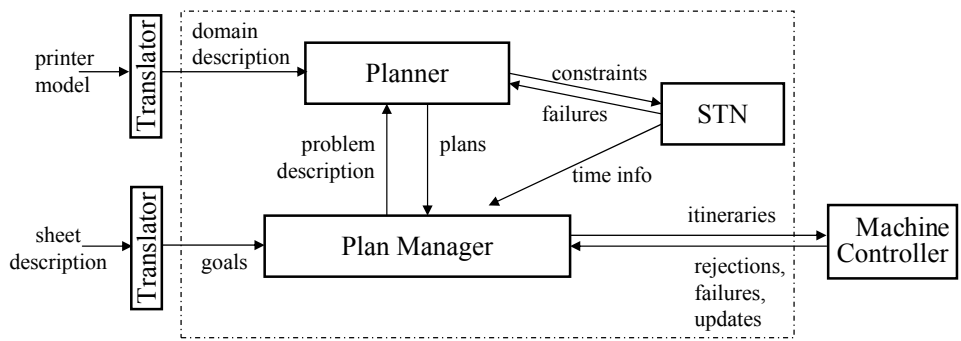

Figure 3: The system architecture, with the planning system indicated by the dashed box.

The planning system must decide how to print all requested sheets as quickly as possible and thus it must determine a plan and schedule for each sheet such that the end time $T$ of the plan that finishes last is minimized. In other words, the planner attempts to minimize the makespan of the combined global plan for all sheets, in essence optimizing the system's overall throughput. Typically there are many feasible plans for any given sheet request; the problem is to quickly find one that minimizes $T$. The optimal plan for a sheet depends not only on the sheet request, but also on the resource commitments present in previously-planned sheets. Any legal series of actions can always be easily scheduled by pushing it far into the future, when the entire printer has become completely idle, but of course this is not desirable. This is an on-line task because the set of sheets grows as time passes and plan execution (i.e., printing) interleaves with plan creation. In fact, because it is the real-world wall clock end time that we want to minimize and because the production of a sheet cannot start until it is planned, the speed of the planner itself affects the value of a plan! However, the system often runs at full capacity, and thus the planner usually need only plan at the rate at which sheets are completed, which again may be several per second. While challenging, the domain is also forgiving: feasible schedules can be found quickly, sub-optimal plans are acceptable, and plan execution is relatively reliable.

A printer controller works in an on-line real-time and continual planning environment with three on-going processes: 1) on-line arrival of new goals; 2) planning for known goals; and 3) execution of previously synthesized plans. Figure 3 shows the inputs and outputs of the planning system, with the domain model and sheet requests entering from the left and communication with the low-level control system on the right. The plan manager is responsible for tracking the status of each goal and invoking the planner when necessary. While planning and execution occur sequentially for any given sheet, these processes will usually be interleaved between different sheets. Figure 4 sketches the different steps in the sheet-plan life cycle managed by the plan manager. Specifically, upon receiving, sheets are put in the *unplanned* first-in-first-out queue (sheets 6 and 7). The sheet planner then picks one sheet at the time from the unplanned queue and tries to find a route-plan for that sheet (sheet 5). Any plan when found is put in the queue of plans that haven't yet been sent to the printer controller (sheets 3 and 4). Another plan manager process regularly checks the planned queue to decide if the earliest starting time of any plan in that queue is close enough to the current wall-clock time and send those plans to the printer controller for execution (sheets 1 and 2). Note that in the figure, time advances downward so plans starting earlier are higher in the figure. Sheets 1, 2, and 3 finish in order; sheets 4 and 5 belongs to a different job and can be scheduled to run concurrently.

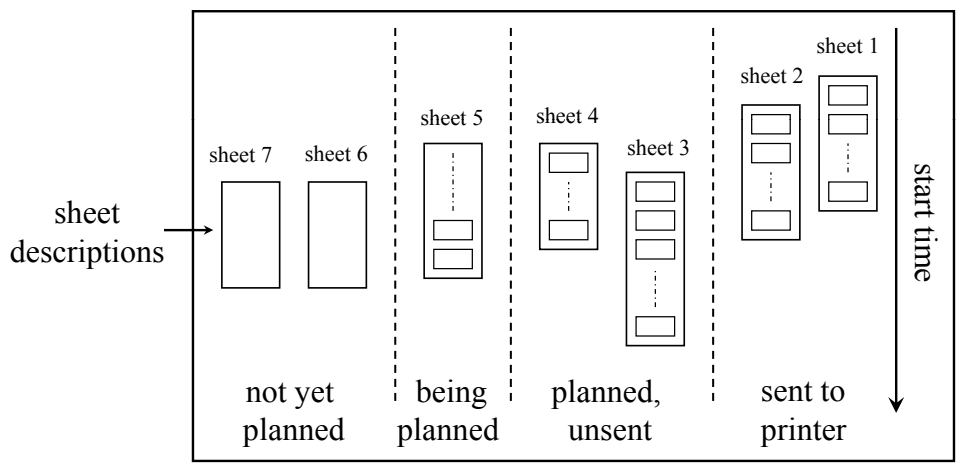

Figure 4: Stages in the life of a sheet in the planning system.

In our application, there is an additional negotiation step after a plan is issued by the planning system and before the plan is committed. First, each plan step is *proposed* by the machine controller to the modules involved. If the individual hardware modules from all steps *accept* their proposed actions, then the plan is *committed*. As we discuss below, this commitment means that modules become responsible for notifying the controller if they fail to complete an action or realize that they will not be able to perform a planned action in the future. After a plan is confirmed, the planner cannot modify it. There is thus some benefit in releasing plans to the machine controller only when their start times approach. If not all modules confirm, then the machine controller notifies the planning system that the proposed plan has been rejected, and the system must produce a new plan. This negotiation process is one reason that we must find a complete plan before starting execution.

Each module has a limited number of discrete actions it can perform, each transforming a sheet in a known deterministic way. For many of these actions, the planner is allowed to control its duration within a range spanning three orders of magnitude (milliseconds to seconds). For example, the planner may choose to transport a sheet faster or slower through a module in order to avoid collisions. Actions may not split a sheet into two pieces or join multiple sheets from different paths in the printer together. This means that a single printed sheet must be created from a single blank sheet of the same size, thereby conflating sheets with material and allowing plans to be a linear sequence of actions. In our domain, adjacent actions must meet in time; sheets cannot be left lingering inside a printer after an action has completed but must immediately begin being transported to its next location.

Sheets are grouped into print jobs. A job is an ordered set of sheets, all of which must eventually arrive at the same destination in the same order in which they were submitted. Multiple jobs may be in production simultaneously, although because sheets from different jobs are not allowed to interleave at a single destination, the number of concurrent jobs is limited by the number of destinations (i.e., finisher trays).

Currently, Xerox uses a constraint-based scheduler to control its high-end and mid-range printers (Fromherz et al., 1999). The scheduler enumerates all possible plans when the machine starts up and stores them in a database. When printing requests arrive on-line, the scheduler picks the first feasible plan from the database and uses temporal constraint processing to schedule its actions. This decoupling of planning and scheduling is insufficient for complex machines for two reasons. First, the number of possible plans is too large to generate ahead of time, and indeed becomes infinite if loops are present, as in the printer shown in Figure 2. Second, the precompiled plans can be poor choices given the existing sheets in the system. For example, sheets should be fed from different feeders depending on when the previous sheets were fed, how large they are, and how long they will dwell in the print engines (which can be a function of sheet thickness and material). For high performance, we must integrate planning and scheduling in an on-line fashion.

Occasionally a module will break down, failing to perform its committed action. Modules can also take themselves off-line intentionally, for example to perform internal re-calibration or diagnosis. Modules may be added or subtracted from the system and this information is passed from the machine controller to the planning system at the right side of Figure 3. The vision of RMP is that the system should provide the highest possible level of productivity that is safe, including running for long periods with degraded capabilities.[1] Meeting this mandate in the context of highly modular systems means that precomputing a limited set of canonical plans and limiting on-line computation to scheduling only is not desirable. For a large system of 200 modules, there are infeasibly many possible degraded configurations to consider. Depending on the capabilities of the machines, the number of possible sheet requests may also make plan precomputation infeasible. Furthermore, even the best pre-computed plan for a given sheet may be suboptimal given the current resource commitments in the printing machine.

To summarize, our domain is finite-state, fully-observable, and specifies classical goals of achievement. However, planning is on-line with additional goals arriving asynchronously. Actions have real-valued durations and use resources. Plans for new goals must respect resource allocations of previous plans. Execution failures and domain model changes can occur on-line, but are rare.

## 3. System Overview

A complete printing system encompasses many components, including print-job submission, print-job management and planning, sheet management and planning, image rendering and distribution, low-level module control, media handling hardware, and exception handling. This paper focuses on planning issues at the sheet level, including exception handling. Before discussing any one issue in great detail, this section provides an overview of those topics that involve sheet planning most directly, including hardware control and exception handling.

---

1. For example, for the safety of the operator, the system should not continue to use a module whose access cover has been opened, even if it were hypothetically possible to repair one portion of the module while another is in use.

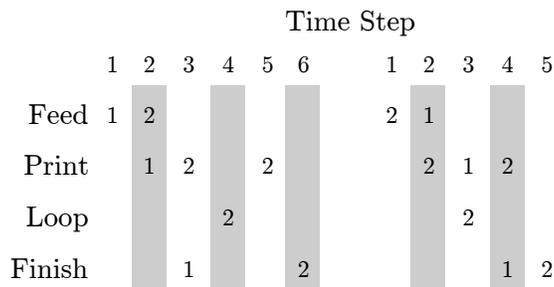

Figure 5: Two different schedules for printing a duplex sheet (2) after a simplex sheet (1):
launching the sheets out of order improves throughput.

Figure 3 shows the basic architecture of the planning system and how it communicates
with the machine controller. The overall objective is to minimize the makespan of the
combined global plan for all sheets, in essence optimizing the system's throughput. We
approximate this by planning *one sheet at a time*, with the objective of having that sheet
finish as quickly as possible while respecting any ordering constraints it may have with other
sheets. Sheets are optimally planned on an individual basis, in order of arrival, without
reconsidering the plans selected for previous sheets. In the figure, the plan manager calls
the planner for each sheet and records the resulting plan. To mitigate the restrictiveness of
this greedy scheme, we represent action times using temporal constraints instead of absolute
times. These constraints are stored in a simple temporal network (Dechter, Meiri, & Pearl,
1991), marked STN in the figure. By maintaining temporal flexibility as long as possible,
we can shift plans for older sheets later in time to make room for starting a new sheet earlier
if that improves overall machine throughput. While this may sound like a rare case, it can
be quite common. Figure 5 illustrates how, for a simplex (single-sided) cover sheet followed
by a duplex (double-sided) sheet, it can be faster overall to launch the second sheet first.

Although this basic architecture is specifically adapted to our on-line setting, the plan-
ner uses no domain-dependent search control knowledge. Furthermore, this mix of goal-
decomposable planning with cross-goal resource constraints is quite common, and we believe
our framework can be useful in any AI system that needs to interleave real-time decision
making, planning, and execution, such as robot operations.

## 3.1 Planning

We have implemented our own temporal planner using an architecture that is adapted to
this on-line domain. As we will see below, the large number of potential plans for a given
sheet and the close interaction between plans and their schedules means that it is much
better to process scheduling constraints during the planning process and allow them to focus
planning on actions that can be executed soon. The planner uses state-space regression,
with temporal information stored in the STN. The STN records a feasible interval for each
time point in each plan. Time points are restricted to occur at specific single times only
when the posted constraints demand it. Because the planner maintains the partial orders
between different actions in plans for different sheets through the STN while conducting

**On-linePlanner**

1. plan the next sheet
2. if an unsent plan starts soon, then
3.    foreach plan, from the oldest through the imminent one
4.       clamp its time points to the earliest possible times
5.       release the plan to the machine controller

**PlanSheet**

6. search queue ← {final state}
7. loop:
8.    dequeue the most promising node
9.    if it is the initial state, then return
10.   foreach applicable action
11.      apply the action
12.      add temporal constraints
13.      foreach potential resource conflict
14.         generate all orderings of the conflicting actions
15.   enqueue any feasible child nodes

Figure 6: Outline of the hybrid planner

the backward state-space search, it can be seen as a hybrid between state-space search and partial-order planning. A sketch of the planner is given in Figure 6. The outer loop corresponds to the plan manager in Figure 3.

After planning a new sheet, the outer loop checks the queue of planned sheets to see if any of them begin soon (step 2). It is imperative to recheck this queue on a periodic basis, so 'soon' is defined to be before some constant amount after the current time and we assume that the time to plan the next sheet will be smaller than this constant. The value of this constant depends on the domain specifics such as communication delay or module preparation time and is currently selected manually. If this assumption is violated, we can interrupt planning the next sheet and start over later. As plans are released and executed, resource contention will only decrease, so the time to plan the new sheet should decrease as well. It is important that new temporal constraints are added by the outer loop only between the planning of individual sheets, as propagation can affect feasible sheet end times and thus could invalidate previously computed search node evaluations if planning were underway.

While maintaining partial orderings between actions seems necessary to mitigate our one-sheet-at-a-time greedy strategy, the planning for individual sheets need not necessarily take the form of state-space regression. We have considered a forward search strategy, such as employed by many modern planners such as FF (Hoffmann & Nebel, 2001) or LAMA (Richter, Helmert, & Westphal, 2008). Initial investigation and preliminary empirical comparisons showed that while a progression planner is easier to implement and easier to extend to handle additional domain complexities, the performance of the regression planner (using the same heuristic) is significantly better in many problems in this domain. This seems to be due mainly to the temporal constraint enforcing that a given sheet should end after the end time of all the previous sheets in the same batch. This constraint interacts well with searching backward from the goal, immediately constraining the end time of the plan. Together with the constraint that actions must abut in time, many possible orderings for resolving resource contention are immediately ruled out. For example, the current sheet cannot be transported to its destination before the previous sheet in the same batch. In addition, some orderings may immediately push the end time of the plan even later, further informing the node evaluation function.

A planner that searches in the forward direction benefits slightly from avoiding logical states that are unreachable from the initial state. However, without a similar temporal constraint for the first action in the plan, few resource allocation orderings can be pruned and the branching due to resource contention increases in direct proportion to the number of plans for previous sheets maintained in the plan manager. Furthermore, the end time of the plan rarely changes until far into the planning processes, making the heuristic less useful. In short, for the first sheet, the performance of forward or backward planners are similar, while as the number of plans managed by the plan manager increases, the backward planner seems to perform better.

Due to details of the machine controller software, the planner must release plans to the machine controller in the same order in which the sheet requests were submitted. This means that sheets submitted before any imminent sheet must be released along with it (step 3). Only at this stage are the allowable intervals of the sheet's time points forcibly reduced to specific absolute times (step 4). Sensibly enough, we ask that each point occur exactly at the earliest possible time. Because the temporal network uses a complete algorithm to maintain the allowable window for each time point (a variation on Cervoni, Cesta, & Oddi, 1994), we are guaranteed that the propagation caused by this temporal clamping process will not introduce any inconsistencies. The clamping happens before plans are issued; thus we do not face the on-line dispatchability problem of Muscettola, Morris, and Tsamardinos (1998).

In our current on-line setting, even though we plan for multiple sheets belonging to different jobs, we build plans for a single sheet at the time. Even if there are many submitted sheets waiting to be planned, this strategy is reasonable given that sheets arrived in sequence and, until the arrival of the last sheet, we do not know how many sheets are in each job and when the planner will receive the individual sheet specifications. Waiting until all sheets are known is impractical as many production jobs, such as billing and payroll, involve jobs with many thousands of sheets that can run for multiple days.

## 3.2 Control

As shown in Figure 2, our system consists of two feeders on the left, two finishing trays on the right, and four print engines with one in each of the four 'quadrants' of the printer. There are three high-speed sheet 'highways' that connect the feeders with the finisher trays. Sheets traveling on the top and bottom highways can be routed to and from the print engines through the 'on-ramps' and 'off-ramps.' For increased modularity, the highways and the on- and off-ramps are made up of only two types of modules: 'straight-through' modules and 'three-way' modules. Each module has its own processor: a Texas Instruments F2811

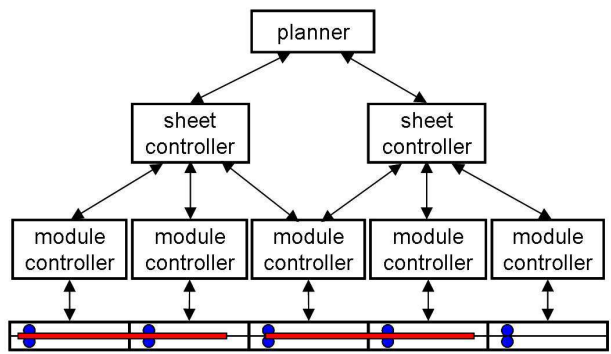

Figure 7: The control system architecture.

DSP. All modules run the same distributed algorithms for state estimation and control and communicate with each other via five controller-area network (CAN) buses, plus a dedicated data-logging bus for debugging purposes. Modules from the same 'quadrant' of the printer reside on the same CAN bus, except for the four print engines, which are on a separate bus. Sheets are moved by roller actuators, called *nips*, that are driven by independently-controlled stepper motors. For sensory feedback, each nip is equipped with an edge-detection sensor on both sides of the nip. In each three-way module, there are three solenoids that drive flipper actuators to direct the sheet along different paths.

Figure 7 shows the control system architecture, which implements a hierarchical approach to distributed plan execution in which a sheet controller manages those module controllers that are currently, or will soon be, in contact with the sheet. Thus the sheet controller group membership is dynamic over the life cycle of a sheet, starting from the feeder all the way to the finisher tray. As soon as a new sheet is sent to the machine controller, a corresponding sheet controller is created that resides on a centralized processor, even though all the module controllers it manages reside locally on the modules themselves. Note that a module controller may be processing commands from multiple sheet controllers, as is the case of the module controller in the middle of Figure 7. While still in contact with the first sheet, it will soon be in contact with the second sheet.

Because a sheet can span multiple modules in our printer, different nips acting on the same sheet must be tightly synchronized in order to avoid damaging or jamming the sheet. However, achieving exact synchronization over a network with uncertain communication delays and stringent bandwidth can be challenging. Moreover, one must consider the limited computation power of a module in the design of such a synchronization scheme. For example, our controller sample time is set at 2 milliseconds. Thus, anything that takes longer than 2 milliseconds to compute within a sampling period will not work. To eliminate the effects of uncertain network delays, the control system uses a delay equalizer, which buffers all sensory feedback messages until an 'apply' time, to make sure that any sensor information is used at the same time by all members of a group of module controllers for the same sheet. To save bandwidth needed for synchronization, each controller uses internal models (or estimators) to keep track of the states of the other controllers on the network in order to limit the need for communications (Crawford, Hindi, Zhou, & Larner, 2009; Hindi, Crawford, & Fromherz, 2005).

The limited network bandwidth has fundamental impact on our choice of the control algorithm. Initially, a linear-quadratic-gaussian (LQG) (Franklin, Powell, & Workman, 1997) controller was used, which has the nice property that its solution constitutes a linear dynamic feedback control law that is easily computed. However, the bandwidth requirements of an LQG controller, which necessitates that more than a dozen way points per sheet be sent over the network, has prompted the adoption of a different kind of controller based on proximate time optimal servo (PTOS) (Hindi, Crawford, Zhou, & Eldershaw, 2008; Franklin et al., 1997), which consumes much less bandwidth. For comparison, a PTOS controller reduces the number of intermediate way points from more than a dozen to two per sheet. Since PTOS is based on time optimal control that uses either the maximum acceleration or deceleration to reach the target control state, this also maximizes the temporal flexibility of the planning actions that our planner can use, thus improving on the overall throughput of the printer.

## 3.3 Previous Work

There has been much interest in the last 15 years in the integration of planning and scheduling techniques. HSTS (Muscettola, 1994) and IxTeT (Ghallab & Laruelle, 1994) are examples of systems that not only select and order the actions necessary to reach a goal, but also specify precise execution times for the actions. The Visopt ShopFloor system of Barták (2002) uses a constraint logic programming approach to incorporate aspects of planning into scheduling. And the Europa system of Frank and Jónsson (2003) uses an novel representation based on attributes and intervals. All of these system use domain representations quite different from the mainstream PDDL language (Fox & Long, 2003) used in planning research and all of them were designed for off-line use, rather than controlling a system during continual execution.

There is currently great interest in extending planning and scheduling techniques to handle more of the complexities found in real industrial applications. For example, PDDL has been extended to handle continuous quantities and durative actions. There are additional dimensions to planning complexity besides expressivity, however. Our work complements the trend in current planning research to extend expressiveness by focusing on the middle ground between planning and scheduling. The domain semantics for printing are more complex than in job shop scheduling but simpler in many ways than PDDL2.1. Choice of actions to perform is important in our domain, but managing resource conflicts is equally important. As in classical scheduling, resource constraints are essential because the printer modules often cannot perform multiple actions at once. But action selection and sequencing are also required because a given sheet can usually be achieved using several different sequences of actions.

Our domain formalization lies between partial-order scheduling and temporal PDDL. Because the optimal actions needed to fulfill any given print-job request may vary depending on the other sheets in the machine, the sequence of actions is not predetermined and classical scheduling formulations such as job-shop scheduling or resource-constrained project scheduling are not expressive enough. This domain clearly subsumes job-shop and flow-shop scheduling: precedence constraints can be encoded by unique preconditions and effects. Open shop scheduling, in which one can choose the order of a predetermined set of actions for each job, does not capture the notion of alternative sequences of actions and is thus also too limited. The positive planning theories of Palacios and Geffner (2002) allow actions to have real-valued durations and to allocate resources, but they cannot delete atoms. This means that they cannot capture even simple transformations like movement that are fundamental in our domain. In fact, optimal plans in our domain may even involve executing the same action multiple times, something that is always unnecessary in a purely positive domain. However, the numeric effects and full durative action generality of PDDL2.1 are not necessary.

Because of the on-line nature of the task and the unambiguous objective function, there is an additional trade-off in this domain between planning time and execution time that is absent from much prior work in planning and scheduling. In our setting the set of sheets is only revealed incrementally over time, unlike in classical temporal planning where the entire problem instance is available at once. And in contrast to much work on continual planning (desJardins, Durfee, Ortiz, & Wolverton, 1999), the tight constraints of our domain require that we produce a complete plan for each sheet before its execution can begin. Our domain emphasizes on-line decision making, which has received only limited attention to date. Our objective is to complete the known print jobs as soon as possible, so taking too long to synthesize a slightly shorter plan is worse than quickly finding a near-optimal solution. This is especially true when rerouting in-flight sheets during exception handling.

Although we present our system as a temporal planner, it fits easily into the tradition of constraint-based scheduling (Smith & Cheng, 1993; Policella, Cesta, Oddi, & Smith, 2007). The main difference is that actions' time points and resource allocations are added incrementally rather than all being present at the start of the search process. The central process of identifying temporal conflicts, posting constraints to resolve them, and computing bounds to guide the search remains the same. In our approach, we attempt to maintain a conflict-free schedule rather than allowing contention to accumulate and then carefully choosing which conflicts to resolve first. Our approach is perhaps similar in spirit to that taken by the IxTeT system (Ghallab & Laruelle, 1994).

Our basic approach of coordinating separate state-space searches via temporal constraints may well be suitable for other on-line planning domains. By planning for individual print jobs and managing multiple plans at the same time, our strategy is similar in spirit to planners that partition goals into subgoals and later merge plans for individual goals (Wah & Chen, 2003; Koehler & Hoffmann, 2000). In our framework, even though each print job is planned locally, the plan manager along with the global temporal database ensures that there are no temporal or resource inconsistencies at any step of the search. It would be interesting to see if the same strategy could be used to solve partitionable STRIPS planning problems effectively.

## 4. Nominal Sheet Planning

The sheet planner builds a plan for each sheet of a job using a combination of regression state-space planning and partial-order scheduling. It plans by adding one module action at a time, starting from a finisher until the sequence of actions reaches a feeder. Adding an action to a sheet's itinerary (i.e., plan) causes resource allocations to be made on any resources required for the execution of that action. Given the media path redundancies in RMP, the planner usually faces multiple choices about which action to add at each planning step. To organize this search, the planner uses best-first A* search with a planning-graph heuristic, adjusted with resource conflicts, that estimates how promising each plan suffix is. Unlike traditional regression planners, to maintain maximum flexibility, all action times such as the start and end of each action and each resource allocation are represented as flexible time points instead of absolute times. Temporal constraints are used to represent the durations of actions and to resolve resource contention by imposing orderings among actions. The planner attempts to minimize the makespan of the combined global plan for all sheets, in essence optimizing the system's throughput. The planner uses no domain-dependent search control knowledge, allowing us to use the same planner to run very different printing systems at full productivity.

## 4.1 Domain Language

We used a two-tiered approach to represent the RMP domain. At the highest level, we use a specialized language that makes it easier for Xerox engineers to model their printers. This language specifies printer configurations as components that are connected to each other. Basic components can have different capabilities and components can be grouped in a hierarchical structure. The model files in this format are then automatically translated into a variation of PDDL2.1, which is then fed into our planner. The automatic translation process instantiates the primitive modules and then converts each module's capabilities into durative actions. The movement of a sheet and the marking actions can be directly translated from the printer model into traditional logical preconditions and effects that test and modify attributes of the sheet. Following the spirit of compositionality of earlier work (Fromherz et al., 1999), the model of the system can be automatically synthesized from models of the individual components.

As in PDDL, we distinguish between two types of input to the planner. Before planning begins, a domain description containing predicate and action templates is provided. Then the problem descriptions arrive on-line, containing initial and goal states, which are sets of literals describing the starting and desired configurations. Our action representation is similar to the durative actions in PDDL2.1, with the notable difference that we use explicit representation of resources. Actions can specify the exclusive use of different types of resources for time intervals specified relative to the action's start or end time. Executing one action may involve allocating multiple resources of various types such as: *unit-capacity*, *multi-capacity*, *cyclic*, and *state* resources. Our actions also specify real-valued duration bounds. That is, one can specify upper and lower bounds and then let the planner choose the desired duration of the action. This is critical to modeling controllable-speed paper paths, which can be very useful in practice. While PDDL allows the specification of duration ranges, we are not aware of any IPC benchmark that does so or any general-purpose planner that supports it.

To summarize, the core part of a domain file is a set of actions, each of which corresponds to a capability of some component in a printer and is a 4-tuple $a = \langle Pre, Eff, dur, Alloc \rangle$, where:

| PrintSimplexAndInvert(?sheet, ?side, ?color, ?image) | |
|---|---|
| preconditions: | Location(?sheet, Printer1-Input) |
| | Blank(?sheet) |
| | SideUp(?sheet,?side) |
| | Opposite(?side, ?other-side) |
| | CanPrint(MarkingEngine, ?color) |
| effects: | Location(?sheet, Printer1-Output) |
| | ¬Location(?sheet, Printer1-Input) |
| | HasImage(?sheet,?side,?image) |
| | ¬Blank(?sheet) |
| | ¬SideUp(?sheet, ?side) |
| | SideUp(?sheet,?other-side) |
| duration: | [13.2 seconds, 15.0 seconds] |
| set-up time: | 0.1 second |
| allocations: | MarkingEngine at ?start + 5.9 for 3.7 seconds |

Figure 8: A simple action specification.

- *Pre* and *Eff* are sets of literals representing the action's preconditions and effects.

- *dur* is a pair $\langle lower, upper \rangle$ of scalars representing the upper and lower bounds on action duration.

- *Alloc* is a set of triplets $\langle res, offset, dur \rangle$ indicating that action $a$ starting at time $s_a$ uses resource *res* during an interval $[s_a + offset, s_a + offset + dur]$. The constraints on different types of resources are:

  - *Unit-capacity:* this type of resource is non-sharable and thus all allocations for a given resource of this type should not overlap. This provides a good model of physical space and is the most common type of resource used in our planner.

  - *Cyclic:* cyclic resource is one special type of unit-capacity resources for which there are repeated durations during which the resources are unavailable for the actions selected by the planner. For example, the unavailable durations may represent routine automatic maintenance of some modules.

  - *Multi-capacity:* there is an upper-bound on the maximum number of allocations for a given resource of this type that can overlap. Moreover, allocations follow a first-in-first-out order. Thus, if there are two allocations $A_1 = [s_{A_1}, e_{A_1}]$ and $A_2 = [s_{A_2}, e_{A_2}]$ then $s_{A_1} \prec s_{A_2}$ implies $e_{A_1} \prec e_{A_2}$.

  - *State resource:* The resource can be labeled using one of a set of 'states'. Allocations for a resource of this type can overlap if and only if they require the resource to be in the same 'state'.

A simple example is given in Figure 8. Set-up time refers to the required time between when an action is committed and its execution begins—certain actions can require extensive preparation on the part of the module before the sheet arrives and the action is really

| | |
|---|---|
| background: | Sheet-23 |
| initial: | Location(Sheet-23, Some-Feeder) |
| | Blank(Sheet-23) |
| | SideUp(Sheet-23,Side 1) |
| goal: | Location(Sheet-23, Upper-Finisher) |
| | HasImage(Sheet-23, Side 1, Image 1) |
| | HasImage(Sheet-23, Side 2, Image 2) |
| | Color(Sheet-23, Side 1, Color) |
| | Color(Sheet-23, Side 2, Black & White) |
| print job ID: | 5 |

Figure 9: A sample sheet specification.

'performed'. For resource usage, the *PrintSimplexAndInvert* action in Figure 8 specifies exclusive use of the *MarkingEngine* from 5.9 seconds after the start of the action until 3.7 seconds later. Printer modules with multiple independent resources or with actions that have short allocation durations relative to the overall action duration can work on multiple sheets simultaneously. In PDDL, arbitrary predicates can be made to hold at the start, end, or over the duration of an action. This expressivity is not needed in our domain and thus we can assume a simple semantics similar to that using in the TGP planner of Smith and Weld (1999) in which: (1) delete effects happen 'at start'; (2) add effects happen 'at end'; (3) preconditions that are deleted are 'at start'; and (4) preconditions that are not deleted are 'over all'. In addition to sheet-dependent literals, sometimes it is convenient to specify actions using preconditions that refer to literals that are independent of the particular goals being sought. This 'background knowledge' about the domain is supplied separately in the machine specification, although it could also be compiled into the action specifications. In our example, the possible colors that engines can put on a sheet of paper (e.g., Black&White, Color, Custom Color) or default sides of papers (e.g., Front, Back) are specified in this way. They are represented similarly to the 'constant' concept in PDDL.

In addition to the static domain description, the on-line sheet requests are modeled by initial and goal state pairs describing the starting and desired sheet configurations. Each new initial/goal pair defines a new object (the sheet) and the associated literals for the planner to track. Specifically, a problem description for a particular sheet is a 4-tuple of ⟨*Job, Initial, Goal, Background*⟩, where *Job* is the id of the print job that that sheet belongs to and *Initial*, *Goal*, and *Background* are sets of literals.

A simple example sheet specification is given in Figure 9. In this example, `Some-Feeder` is a virtual location where all sheet sources are placed and `Upper-Finisher` is one particular finisher where all sheets that belong to print job 5 need to be routed to. In terms of Figure 3, the feeder location literal is added to the goal by the plan manager, which maintains a table of active jobs and the finisher assigned to each. Finisher assignment is handled by extracting the finisher selected by the planner in its plan for the first sheet of the job. Because finishing requires actions in the plan and actions are never reconsidered (only rescheduled), the planner can never reconsider a job's finisher assignment, even if it hasn't begun production yet.

Given a domain description (top left of Figure 3) and a low-level delay constant $t_{delay}$ capturing the latency of the machine controller software, the planner then accepts a stream of sheets arriving asynchronously over time. Note that sheets may belong to different print jobs being printed in parallel; within their print job, sheets need to be routed to the same finisher (among multiple finishers) and finish in order. This stream corresponds to the standard notion of a PDDL problem instance. For each sheet, the planner must eventually return a plan: a sequence of actions labeled with start times (in absolute wall clock time) that will transform the initial state into the goal state. Any allocations made on the same unit-capacity resource by multiple actions must not overlap in time (state and multi-capacity resources have different constraints as described earlier). Happily, plans for individual sheets are independent except for these interactions through resources. Additional constraints on the planner include:

- plans for sheets with the same print job id must finish at the same destination,

- plans for sheets with the same print job id must finish in the same order in which the jobs were submitted,

- the first action in each plan must not begin sooner than $t_{delay}$ seconds after it is issued by the planner (with $t_{delay}$ represents the delays in communication and negotiation with the printer module controller),

- subsequent actions must begin at times that obey the duration constraints specified for the previous action (thus it is assumed that the previous action ends just as the next action starts).

## 4.2 Temporal Reasoning

Printer control is a rich temporal domain with real-time constraints: (i) between wall-clock time and the plans for individual sheets, (2) between plans for different sheets, and (3) between the planner and the machine controller. Thus, fast temporal constraint propagation, consistency checking, and querying are extremely important in our planner. We maintain the temporal constraints using a Simple Temporal Network (STN) (Dechter et al., 1991), represented by the box named *STN* in Figure 3. Essentially, the network contains a set of temporal time points $t_i$ and constraints between them of the form $l_b \leq t_i - t_j \leq u_b$. The time points managed by the STN include action start and end times and resource allocation start and end times. Temporal constraints maintained in the STN are:

- constraints on wall-clock action start time;

- action start and end times should be within the action duration range;

- constraints between action start time and resource allocation by that action; and

- conflicts for various types of resources.

Because we use an A* search strategy that maintains multiple open search nodes, there is a separate STN for each node. Temporal constraints are added to the appropriate STN when a search node is expanded. Whenever a new constraint is added, propagation tightens the upper and lower bounds on the domain of each affected time point. While this can lead to more memory usage and extra overhead, it allows us not having to deal with temporal constraint retraction, which is needed if a single STN is used for multiple search nodes. Retracting temporal constraints from an STN is a complicated and time consuming process. Because the planner must run indefinitely, we perform garbage collection on time points in the STN between sheet planning episodes, harvesting those that lie in the past.

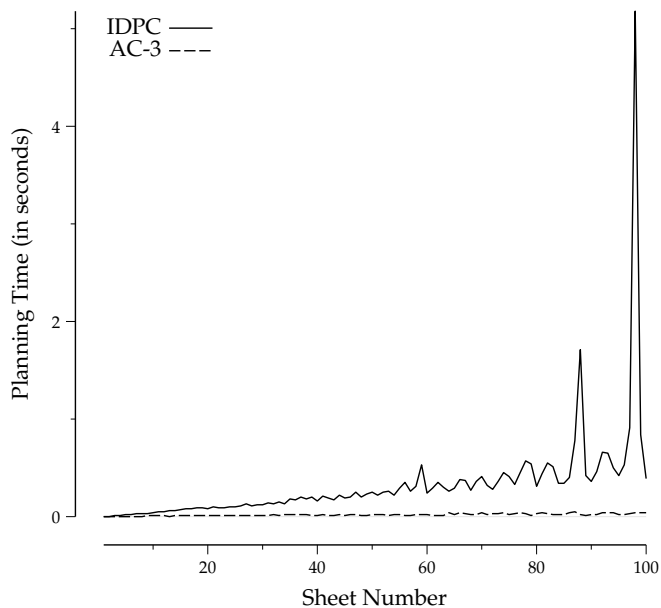

Figure 10: Simple arc consistency is faster than incremental directed path consistency for maintaining our STNs.

All time points are flexible until the plans they belong to are sent to the machine controller. After planning a new sheet, the plan manager checks the queue of planned sheets to see if there are any that could begin soon. If there are, those plans are released to the machine controller to execute. New temporal constraints are added that freeze the start and end times of actions belonging to plans sent to the controller. Those time points are frozen at the earliest possible wall-clock time as indicated by the STN based on its constraint set. Those constraints can cause significant propagation and in turn (1) freeze the start and end times of resource allocations related to actions in the frozen plans; and (2) tighten the starting times of actions in the remaining plans.

The original representation of an STN as a complete matrix of time relations updated by all-pairs shortest paths (Dechter et al., 1991) is much too inefficient for our purposes. We have implemented two versions of the STN. One uses an incremental directed path consistency (IDPC) algorithm (Chleq, 1995), which may change the values on edges in the constraint graph as well as introduce new edges but requires only linear time to find the minimum and maximum interval between any two time points in the database. The other uses arc consistency (Cervoni et al., 1994) and maintains for each point its minimum and maximum times from $t_0$, the reference time point. In this latter method, one cannot easily obtain the exact relations between arbitrary time points, only their relations with $t_0$. However, as long as inconsistency can be efficiently detected when constraints are added, we do not need to query the relations of arbitrary pairs of points, and the efficiency gains are welcome. New arcs are never added to the network during propagation and existing ones are not modified, which means that copying the network for a new search node does not entail copying all the arcs. As the Figure 10 attests, this results in dramatic time savings and this technique is used in our current implementation. We further improved our implementation by (1) using change flags to facilitate faster cycle detection for temporal consistency checking and (2) converting all times and durations to integer values (with user defined precision) to eliminate rounding effects and increase speed.

## 4.3 Planning a Sheet

When planning individual sheets, the regressed state representation contains the state of the sheet, which may be only partially specified. A* search is used to find the optimal plan for the current sheet, in the context of all previous sheets. After the optimal plan for a sheet is found, the resource allocations and *STN* used for the plan are passed back to the plan manager and become the basis for planning the next sheet.

One unusual feature of our planning approach is that we seamlessly integrate planning and scheduling. Starting times of actions are not fixed but merely constrained by temporal ordering constraints in the STN. We insist that any potential overlaps in allocations for the same resource be resolved immediately, resulting in potentially multiple children for a single action choice. This allows temporal propagation to update the action time bounds and guide plan search. While the plan for a single sheet is a totally-ordered sequence of actions, there are partial orders between actions that belong to plans of different sheets to represent the resource conflict resolutions.

### 4.3.1 State Representation

Because the plan must be feasible in the context of previous plans, the state contains information both about the current sheet and previous plans. More specifically, the state is a 3-tuple $\langle Literals, Tdb, Rsrcs \rangle$ in which:

**Literals** describes the regressed logical state of the current sheet. We distinguish between literals whose status is true, false, or unknown (Le, Baral, Zhang, & Tran, 2004). The distinction between false and unknown literals is important in our domain because there may be fine-grained restrictions on the acceptable values for unspecified attributes of the sheet. For example, if a sheet $S$ is the first of a given print job, then the finisher representing the final location of the sheet is unknown because it can be any finisher that is not allocated to another print job when the plan for that sheet $S$ is executed. As we discuss below, we allow regression to match unknown literals with both true and false effects of actions; in this sense 'unknown' can function like 'don't care'. In our implementation, we represent explicitly those literals that are currently true and those whose value is unknown, with false literals being represented implicitly.

**Tdb** is the temporal database represented as a simple temporal network (STN) containing all known time points and the current constraints between them. This includes constraints between different plans, between actions in the same plan, as well as against the wall-clock time. Examples of time points include the start/end times of actions or resource allocations. As soon as a plan for a given sheet is sent to the machine (sheets 1 and 2 in Figure 4), time points associated with that plan in the *Tdb* are no longer allowed to float but are clamped at their lower bounds. All other time points are flexible.

**Rsrcs** is the set of current resource allocations, representing the commitments made to plans of previous sheets and the partial plan of the current sheet. Each resource allocation is of the form $\langle res, tp_1, tp_2 \rangle$ with *res* is a particular resource and $tp_1, tp_2$ are two time points in the *Tdb* representing the duration *res* is allocated to some action. Note that there are multiple resources in the domain and each resource can have multiple (overlapping or non-overlapping depending on the resource type) resource allocations. In our implementation, we maintain an ordered list of the allocations on each resource, most recent to oldest.

In essence, the state contains information reflecting the strategy of our planner: hybrid between state-space sequential temporal regression search and partial order scheduling. The *Literals* and the action start and end time-points represent the temporal-planning regressed state and the *Rsrcs* and the temporal orderings between competing resource allocations represent partial-order scheduling constraints between actions in the plans of different sheets.

### 4.3.2 BRANCHING ON APPLICABLE ACTIONS

Each regressed logical state in our planner is a 3-tuple $L = \langle L_t, L_f, L_u \rangle$ where $L_t$, $L_f$, and $L_u$ are the disjoint sets of literals that are true, false, and unknown, respectively. If $Pre^+(a)$ and $Pre^-(a)$ are the sets of positive and negative preconditions and $Add(a)$ and $Del(a)$ are the sets of positive and negative effects of action $a$, then the regression rules used to determine action applicability and update the state literals are:

**Applicability** Action $a$ is applicable to the literal set $L$ if (1) none of its effects are inconsistent with $L$ and (2) any preconditions not modified by the effects of $a$ are consistent with $L$. More formally, (1) $(Add(a) \bigcap L_f = \emptyset) \wedge (Del(a) \bigcap L_t = \emptyset)$, and (2) $(Pre^+(a) \bigcap L_f \subset Del(a)) \wedge (Pre^-(a) \bigcap L_t \subset Add(a))$.

In many planning settings, an additional criterion for applicability can be added without destroying completeness: at least one effect of $a$ must match $L$ $((Add(a) \bigcap L_t \neq \emptyset) \vee (Del(a) \bigcap L_f \neq \emptyset))$. This is not necessarily valid in our setting because adding a 'no-op' action $a$ may give more time for an existing resource allocation to run out, enabling other actions to be used which might lead to a shorter plan.

**Update** The regression of $L = \langle L_t, L_f, L_u \rangle$ over an applicable action $a$ is derived by undoing the effects of $a$ and unioning the result with $a$'s preconditions. For a given literal $l$ modified by an effect of $a$, its status will be unknown in the regressed state unless it is also specified by a corresponding precondition of $a$ (e.g., $\neg l$ is a precondition of $a$).

More formally, (1) $L_t = (L_t \setminus Add(a)) \bigcup Pre^+(a)$; (2) $L_f = (L_f \setminus Del(a)) \bigcup Pre^-(A)$; and 3) $L_u = (L_u \bigcup (Add(a) \bigcup Del(a))) \setminus (Pre^+(a) \bigcup Pre^-(a))$

Given that $|L_f|$ is usually much larger than $|L_t|$ and $|L_u|$ in our domain, we explicitly store $L_t$ and $L_u$ in our current implementation and use the closed-world assumption to imply that all other literals belong to $L_f$. The modeling translator we provide to Xerox engineers for modeling printers encourages all effects to be mentioned in preconditions, reducing the growth of the number of unknown literals. For example, if $x \in Add(a)$ then $\neg x \in Pre^-(a)$.

Although it is not usually the case in our domain, we should note that if the goal state were always fully specified (with no unknown literals) and every action's effects had corresponding preconditions, all regressed states would be fully specified. One could then simplify the logical state representation to $L = \langle L_t, L_f \rangle$ and simplify the regression rules to

**Applicability** Action $a$ is applicable iff all of its effects match in $L$: $Add(a) \subseteq L_t$ and $Del(a) \subseteq L_f$.

**Update** Regressing $\langle L_t, L_f \rangle$ through $a$ gives $\langle (L_t \setminus Add(a)) \bigcup Del(A), (L_f \setminus Del(a)) \bigcup Add(a) \rangle$

A plan is considered complete if its literals unify with the desired initial state (step 9 in Figure 6). After the optimal plan for a sheet is found, the temporal database used for the plan is passed back to the outer loop in Figure 6 and becomes the basis for planning the next sheet. Because feasible windows are maintained around the time points in a plan until the plan is released to the machine controller, subsequent plans are allowed to make earlier allocations on the same resources and push actions in earlier plans later. If such an ordering leads to an earlier end time for the newer goal, it will be selected. This provides a way for a complex job $j_2$ that is submitted after a simple job $j_1$ to start its execution in the printer earlier than $j_1$. Out of order starts are allowed as long as all sheets in each print job finish in the correct order. This can often provide important productivity gains.

### 4.3.3 Branching on Resource Allocation Orderings

While the first step in creating regressed states is to branch over the actions applicable in $L$, applying each candidate action $a$ can in fact result in multiple child nodes due to resource contention. Some scheduling algorithms use complex reasoning over disjunctive constraints to avoid premature branching on ordering decisions that might well be resolved by propagation (Baptiste & Pape, 1995). We take a different approach, insisting that any potential overlaps in allocations for the same resource be resolved immediately. Temporal constraints are posted to order any potentially overlapping allocations and these changes propagate to the action times. Because many action durations are relatively rigid in typical printers, this aggressive commitment can propagate to cause changes in the potential end times of a plan, immediately helping to guide the search process. Because multiple orderings may be possible, there may be many resulting child search nodes.

For example, in Figure 4, assume that $a$ is the current candidate action when searching for a plan for sheet 5 and that $a$ uses resource $r$ for a duration $[s, e]$. We also assume that there are $n$ actions in the plans for sheets 1–4 that also use $r$, implying $n$ existing non-overlapping resource allocations $[s_1, e_1]....[s_n, e_n]$ and corresponding time points in the temporal database. When trying to allocate $r$ for $a$, one obvious and consistent choice is putting it after all other previous allocations by adding the temporal constraint $e_n \prec s$. However, there can also be gaps between the existing allocations $[s_i, e_i]$, allowing us to post constraints such as $e_i \prec s \prec e \prec s_{i+1}$. Each such possible allocation for $r$ generates a distinct child node in the search space. Because action $a$ can use several different resources $r$, the number of branches is potentially quite large. However, immediately resolving any potential overlaps in allocations for the same resource avoids the introduction of disjunctions in the temporal network, maintaining the tractability of temporal constraint propagation.

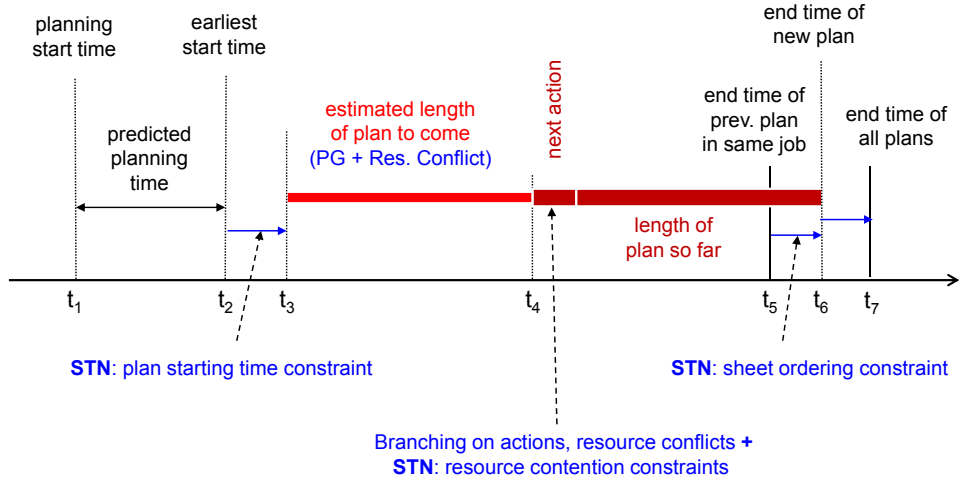

Figure 11: Important time points for constructing and evaluating a plan.

In summary, at every branch in the planner's search space, we modify the logical state by branching over relevant actions and potentially introduce different temporal constraints in order to resolve resource contention. Because each branch results in a different irrevocable choice that is reflected in the final plan, the state at each node in the planner's search tree is unique. Therefore, we do not need to consider the problem of duplicating search effort due to reaching the same state by two different search paths.

### 4.3.4 Heuristic Estimation

For each potential plan suffix, a lower bound is computed on the remaining makespan, in order to guide the planner's A* search. Figure 11 illustrates how this heuristic estimate is used. In the figure, *planning start time* ($t_1$) refers to the wall-clock time at which the planning process started and *earliest-start-time = current wall-clock time + predicted planning time* ($t_2$) is the estimated time at which we will find a plan for the current sheet and thus the earliest time that any action can be scheduled to begin. Note that in practice, the machine controller communication and negotiation time is also added to the predicted planning time. The hypothetical start time of the plan when found ($t_3$) is constrained to happen after this earliest possible wall-clock plan execution time ($t_2 \prec t_3$). A plan is constrained to end after those for previous sheets in the same print job ($t_5 \prec t_6$), but is not necessarily constrained to start after or before plans for previous sheets. The start time of the next action added to the regressed partial plan ($t_4$) is constrained to occur at least $D$ after the hypothetical plan starting time ($t_3 + D \prec t_4$) where $D$ is the heuristic value on the makespan of the remaining plan to complete the current regressed partial-plan.

Our overall objective is to minimize the earliest possible end time of *all* plans, including the sheet that we are planning for. This is indicated by the lower-bound on the floating time point $t_7$ in Figure 11. This time point is constrained to be after the end time points of all the sheets that have been planned and the one currently being planned. For the current sheet, this is represented by the constraints $t_6 \prec t_7$ as shown in Figure 11. Because $t_6$ is constrained to end after the completion time of all the other planned sheets in the same print job, the constraint essentially pushes $t_7$ to be after all the sheets in the current print job end. To support this objective function, the primary criterion evaluating the promise of a partial plan (step 8 in Figure 6) is the estimate of the earliest possible happening time for $t_7$, indicated by the STN embedded in this search node, after all constraints shown in Figure 11 are added in the current branch.

The key duration that affects $t_7$ is the heuristic estimate of the lower bound on the additional makespan required to complete the current regressed plan. This heuristic value is indicated in Figure 11 by *estimated remaining makespan* between $t_3$ and $t_4$. By adding the constraint $t_2 \prec t_3$, the insertion may thus change the earliest time of all the subsequent time points $t_4$, $t_5$, $t_6$ and $t_7$. It may also introduce an inconsistency in the temporal database, in which case we can safely abandon the plan. Given that the current plan should end after the end time of all previous sheets in the same print job ($t_5 \prec t_6$), our objective function is to minimize $t_7$ without causing any inconsistency in the temporal database. We break ties in favor of:

- smaller $t_6$ (e.g., end time of the current print job)

- smaller predicted makespan ($t_6 - t_3$)

- larger currently realized makespan ($t_6 - t_4$). This is analogous to breaking ties on $f(n)$ in A* search with larger $g(n)$, and thus encourages further extension of plans nearer to a goal.

The performance of our search-based planner heavily depends on the quality of the heuristic estimating the makespan-to-go. We estimate $D$ by building the temporal planning graph with adjustment for both logical mutex and resource contentions. For the rest of this section, we will discuss the details of how $D$ is computed. Overall, we want an effective planning heuristic that is:

- *Admissible:* because maintaining high plan quality (high productivity of the printer) is an important criterion for our customer.

- *Informed* and *easy to compute*: because in most cases, we are only allowed a fraction of a second to find a feasible plan. Any delay in finding a plan will delay plan start execution time and thus reduce the overall productivity.

To derive an admissible estimate of the duration required to achieve a given set of goals $G$ from the initial state, we perform dynamic programming over the explicit representation of the bi-level temporal planning graph, which was described in the TGP system (Smith & Weld, 1999). In TGP, the planning graph is represented by a fact level and an action level. Starting with the initial state at time $t = 0$, the graph is grown forward in time with actions being activated when all of their preconditions are satisfied and non-mutex. There are three types of mutual exclusion relations (fact-fact, fact-action, action-action) that are propagated during the graph building process. The graph expansion phase alternates with the plan extraction phase starting from the time point at which all the goals appear non-mutex in the graph.

In our graph expansion algorithm, for each action $a$ and fact $f$, we store the first times $t_a$ and $t_f$ at which $a$ can optimistically occur or $f$ can optimistically be achieved. They correspond to the first times at which $a$ and $f$ appear in the temporal planning graph. For mutex propagation, we also store the first time point at which each pair of facts $\langle f_1, f_2 \rangle$ can be achieved together and each pair of actions $\langle a_1, a_2 \rangle$ can execute together. In the planning graph, those are the first time points that $\langle f_1, f_2 \rangle$ and $\langle a_1, a_2 \rangle$ become non-mutex. In our implementation, a fact-action mutex between fact $f$ and action $a$ is converted into action mutex $\langle noop_f, a \rangle$, as we will discuss later.

1. To begin: $\forall f, a, f_1, f_2, a_1, a_2 : t_a = t_f = t_{\langle f_1, f_2 \rangle} = t_{\langle a_1, a_2 \rangle} = \infty$.

2. Let $I$ be the initial state: $\forall f, f_1, f_2 \in I : t_f = 0, t_{\langle f_1, f_2 \rangle} = 0$.

3. Dynamically update the values of $t_a, t_f, t_{\langle f_1, f_2 \rangle}, t_{\langle a_1, a_2 \rangle}$ starting from the initial state $I$ and time $t = 0$ as follows:

$$t_a = \max \left( setup\_time(a), \max_{f \in Prec(a)} t_f, \max_{f_1, f_2 \in Prec(a)} t_{\langle f_1, f_2 \rangle} \right) \tag{1}$$

$$t_f = \min_{f \in Eff(a)} (t_a + dur(a)) \tag{2}$$

$$t_{\langle f_1, f_2 \rangle} = \min \left( t_{\langle a_1, a_2 \rangle} + \max_{f_1 \in Eff(a_1), f_2 \in Eff(a_2)} (dur(a_1), dur(a_2)) \right) \tag{3}$$

$$t_{\langle a_1, a_2 \rangle} = \max \left( t_{a_1}, t_{a_2}, \max_{f_1 \in Prec(a_1), f_2 \in Prec(a_2)} t_{\langle f_1, f_2 \rangle} \right) \tag{4}$$

The updates are done in the increasing order of time, as usual for planning-graph building algorithms.

4. Stop when $\forall g \in G : t_g < \infty$ and $\forall g_1, g_2 \in G : t_{\langle g_1, g_2 \rangle} < \infty$ or we reach a fixed point.

In the equations (1)-(4) as shown above, the actions include the 'noop' actions as in the normal planning graph. Those actions start from the time point at which a fact is first achieved. The mutex relation between a noop and an action is equivalent to a fact-action mutex as described by Smith and Weld (1999). While the overall plan for all sheets is highly parallel, the plan for a single sheet is sequential. Therefore, we currently use the serial version of the temporal planning graph, which is also faster to build and consumes less memory. In this version, two non-noop actions are always mutex with each other. Therefore, we do not need to store and reason about action mutexes and thus the value of $t_{\langle a_1, a_2 \rangle}$ in eq.(4) is only applicable to mutexes between a noop and a real action. In our implementation, we build the graph starting from $t = 0$ by putting in events of (1) activating an action (updating $t_a$); (2) activating a fact (updating $t_f$); and (3) removing a fact mutex (updating $t_{\langle f_1, f_2 \rangle}$), ordered by the time they occur. Each event will trigger new events to happen at a later time. For example, adding a new fact $f$ or removing a fact mutex $\langle f_1, f_2 \rangle$ can activate actions supported by $f$ or by both $f_1$ and $f_2$, and activating action $a$ will add events of activating facts in $\mathit{Effect}(a)$ and/or removing fact mutexes between $\mathit{Effect}(a)$ and 'noop' (facts) that are not mutex with $\mathit{Precond}(a)$. We also only explicitly store the fact-fact mutex timing values $t_{\langle f_1, f_2 \rangle}$ but none of the action mutexes $t_{\langle a_1, a_2 \rangle}$, instead reasoning about them on-the-fly.

The time at which all the goals are achieved pair-wise non-mutex is the heuristic value estimating the remaining makespan to achieve the goal state (see Figure 11). While most regression planners (Haslum & Geffner, 2001; Nguyen, Kambhampati, & Nigenda, 2002) compute their heuristic once (until a fixed point is reached) before the planning process begins, in our case, the planning graph expansion process may be revisited if goals representing a regressed state do not appear non-mutex in the graph and a fixed point was not reached in the previous round of expansion. Because only pair-wise mutexes are taken into account while building the graph, the estimated value is an underestimate of the makespan of any plan that can achieve the goal. Therefore, the returned value by the planning graph will lead to a underestimate (admissible heuristic) for both our objective function (overall end time $t_7$) and tie breakers (current sheet end time $t_6$ and current estimated makespan $t_6 - t_3$) as described above. Therefore, using this estimate, the planner will return plan $p$ with an optimal end time (minimum $t_7$) and $p$ also has a minimum makespan among all plans with the same end time.

**Incorporating resource mutexes** The planning graph discussed until now assumes that the printer is empty. Thus, we create the planning graph similar to the procedure used in an off-line planner in which we assume the interference relations only occur between actions related to a given sheet that we are planning for. If the machine is empty, the heuristic is generally correct for simple sheets such as simplex printing and nearly correct for complicated sheets such as duplex printing.

However, most of the time, the printer is not empty and there are plans for sheets that are either (1) executing; or (2) found by the planner but haven't been sent to the machine controller yet. Those plans involve resource allocations, either at fixed time points (for (1)) or at still flexible ones (for (2)). To find a more effective heuristic in those scenarios where the machine is busy, we take into account resource mutexes, thus incorporating scheduling resource contention constraints into the temporal planning graph. Figure 12 shows the pseudo-code of the algorithm. The key observation is that, to find the earliest time $t_a$ at which an action $a$ can possibly execute, a necessary condition is not only that all of $a$'s preconditions appear non-mutex in the planning graph but also that there is no resource conflict between any resource $r$ used by $a$ and all current allocations of $r$ (given to previous plans and by external processes.)

As shown in the example action description in Figure 8, each resource allocation of action $a$ is represented as a triple $\langle r, o, d \rangle$. If $a$ starts at $s_a$, this means that resource $r$ is used from $s_a + o$ for a duration $d$, which is normally different from the duration $d_a$ of $a$. For example, the lone resource usage for action $\mathit{PrintSimplexAndInvert}$ in Fig-

1. Resource types: $r_1, r_2, ....r_n$
2. All resource allocations: $\{R_1, R_2, ....R_n\}$
   with $R_i = \{[s_{i1}, e_{i1}], [s_{i2}, e_{i2}], ...[s_{im}, e_{im}]\}$ the ordered list of allocations for $r_i$

**Function** $CheckEarliest(r, t, d)$
3.   $r$: resource
4.   $t$: earliest time intend to use $r$
5.   $d$: duration intend to use $r$
6.   $R = \{[s_1, e_1], [s_2, e_2], ...[s_m, e_m]\}$: current allocations for $r$.
7.   $t_{min}$: earliest time of a non-conflict allocation for $r$; initialize to: $t_{min} \leftarrow t$
8.   **for each** allocation $l = [s_k, e_k] \in R$ check
9.     **if** we can reserve $r$ for a duration of $d$ before $l$ starts: $Latest(s_k) > t_{min} + d$
10.       then go to line 14
11.    **else** move forward to the next possible opening at the end of allocation $a$
12.       $t_{min} \leftarrow Earliest(e_k)$
13.   **end for;**
14.   return($t_{min}$)
**end function;**

Building the temporal planning graph
15.   when consider adding action $a$ to the planning graph
16.   Initialize $t_a$ to the earliest time at which $Prec(a)$ achieved non-mutex (eq.1)
17.   Resource allocations of $a$: $R_a = \{\langle r_1, o_1, d_1 \rangle, \langle r_2, o_2, d_2 \rangle, ..., \langle r_n, o_n, d_n \rangle\}$
18.   **for each** allocation $l = \langle r_k, o_k, d_k \rangle$ check the earliest non-conflict time for $l$
19.     $t_a \leftarrow CheckEarliest(r_i, t_a + o_i, d_i) - o_i$
20.   **end for;**
21.   add action $a$ to the temporal planning graph at $t_a$ until all
      goals appear non pairwise mutex;

Figure 12: Building the temporal planning graph with adjustments for resource conflicts.

ure 8 is $(MarkingEngine, 5.9, 3.7)$. Lines 17-19 in Figure 12 show that when building the graph, for each action $a$ that has all of its preconditions satisfied at $t_a$, the algorithm goes through all resources used by $a$. For each resource allocation $\langle r, o, d \rangle$, it calls the function $CheckEarliest(r, t_a + o, d)$ to update $t_a$, the earliest executable time that $a$ can start without overlapping with any of the previous resource allocations of $r$. The pseudo-code of function $CheckEarliest(r, t, d)$ is self-explanatory in that we try to find the earliest time point after $t$ at which we can 'slot' the allocation for $r$ with duration $d$ in without overlapping with the previous allocations of $r$.

Figure 13 shows one example to demonstrate this algorithm. In this example, we try to find the starting time for action $a$, which needs two unit-capacity resource allocations $\langle r_1, o_1, d_1 \rangle$ and $\langle r_2, o_2, d_2 \rangle$ as shown on the top-left corner. Assume that when building the graph, all of $A$'s preconditions can first be achieved non-mutex at time $t_1$. Referring to Figure 13, the fixed allocations for $r_1$ are allocations (1), (2), and(3) and for $r_2$ are (1) and (2). The flexible allocations, shown with their upper/lower bound constraints are (4) for $r_1$ and (3) and (4) for $r_2$. Starting from $t_1$, the first time point at which we can allocate $r_1$ for a duration $d_1$ without overlapping with previous allocations, both fixed and flexible, is $t_3$. Thus, we adjust the new earliest possible starting time for $a$ to $t_2 = t_3 - o_1 \succ t_1$. Given the new earliest possible starting time $t_a = t_2$, we find that the earliest time point from $t_2$ at which we can allocate resource for $r_2$ is $t_5$. Given that $t_4 = t_5 - o_2 \succ t_2$, we will then take $t_4$ as the final earliest starting time for $a$ and activate action $a$ at $t_4$ in the graph (instead of the original value $t_1$)[2].

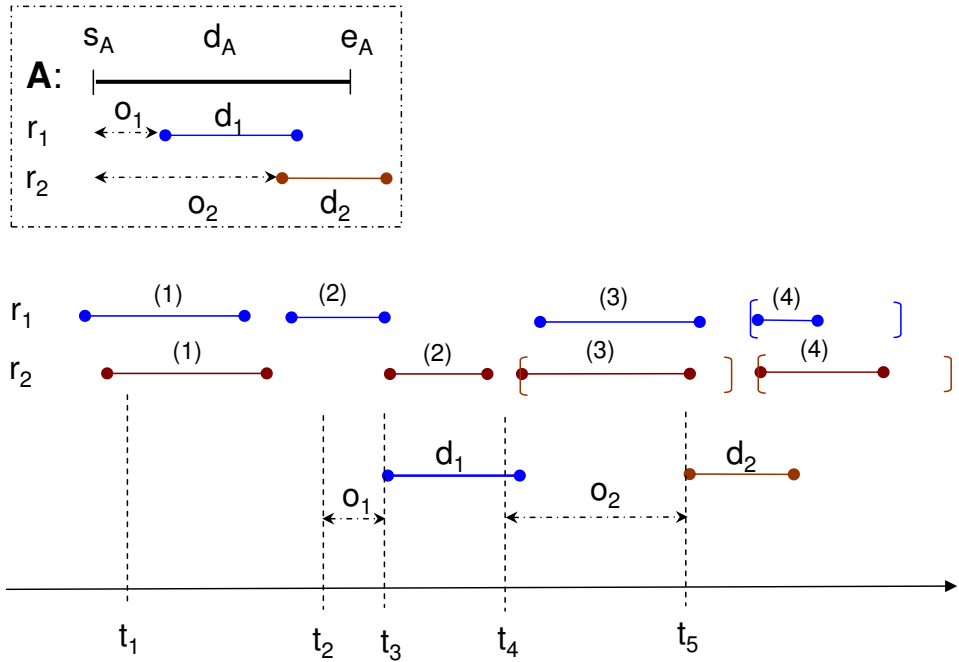

Figure 13: Example of action starting time adjustment using resource contentions

With resource mutexes, the starting times of actions are adjusted to higher than the time points at which their preconditions can be achieved, and thus the time point $t_G$ at which all the goals appear non-mutex in the graph is not an underestimation on the makespan of the remaining plan (value $t_4 - t_3$ in Figure 11). Thus, $t_G$ can be higher than the summation of the durations of actions in the optimal (serial) plan. However, $t_G$ still underestimates the first time that we can achieve the goals and thus is still an admissible heuristic for our main objective function of minimizing the end time of current printing sheets (minimizing $t_7$ in Figure 11). However, if we do not use the resource mutexes, then both the heuristic estimate for end time ($t_7$) and the tie-breaker on plan makespan ($t_6 - t_3$) are admissible while with resource mutexes, only the estimate of $t_7$ is admissible and the tie-breaker $t_6 - t_3$ can be inadmissible. Thus, the A* algorithm is still guaranteed to find an optimal solution, minimizing the plan end-time, but the final solution will not be guaranteed to have a shortest duration among all plans that finish earliest.

---

2. Note that we only go through all resource allocations of a given action one time (lines 18-20 in Figure 12). Therefore, even when action $A$ in Figure 13 is added at $t_4$, there is still a potential conflict for resource $r_1$ consumed by $A$ (with the existing allocation (3) of $r_1$). However, by not repeating the procedure (lines 18-20 in Figure 12) multiple times until a fixed point is reached (and potentially returning better heuristic estimate), we seek the balance between heuristic quality and heuristic computation time.

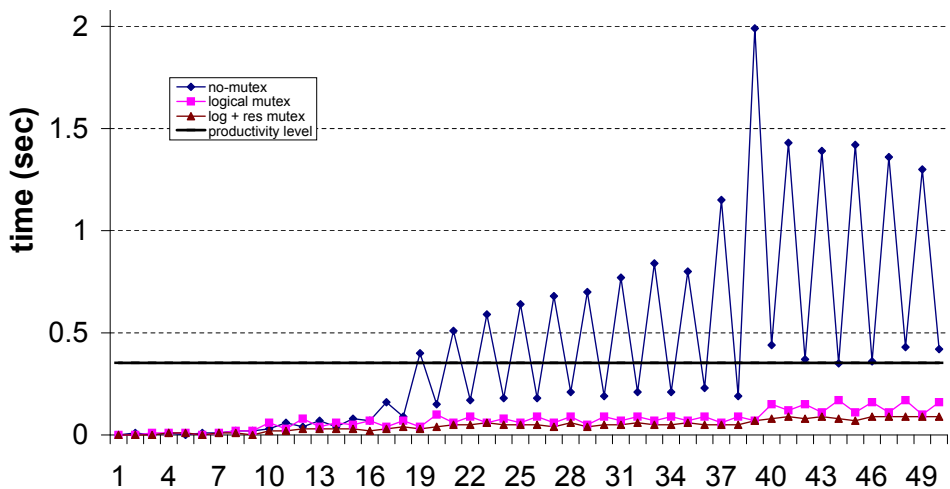

Figure 14: Performance for the prototype built by our industrial partner.

## 4.4 Evaluation of Nominal Planning

In collaboration with Xerox, we have deployed the planner to control three physical prototype multi-engine printers. These deployments have been successful and the planner has also been used in simulation to control hundreds of hypothetical printer configurations. The planner is written in Objective Caml, a dialect of ML, and communicates with the job submitter and the machine controller using ASCII text over sockets. The planner can also communicate with a plan visualizer to graphically display the plans. The first two videos in the on-line appendix show the planner controlling the prototype depicted in Figures 1 and 2 at full productivity, both using the visualizer (video 1, 'nominal simulation') and the hardware (video 2, 'nominal hardware'). A full description of the videos can be found in the textual appendix of the paper. The shortest single plan for the machine has 25 actions. Given that there are many sheets in the printer at any given time and the planner can plan ahead, the plan manager consistently manages dozens of plans and hundreds of actions. During planning, the planner needs to do temporal reasoning regarding the conflict between actions in the current plans and hundreds of actions in previous plans. Even so, the planner consistently on average produces plans within the 0.27 seconds required to keep the printer running at full productivity (220 pages/minute). For one of the most complex current Xerox commercial products, the planner can regularly find the optimal plan within 0.01 seconds and can plan ahead hundreds of sheets. The ability to use domain-independent planning techniques allows us to use the same planner for very different configurations, without needing any hand-tuned control rules. For the rest of this section, we will elaborate on these results.

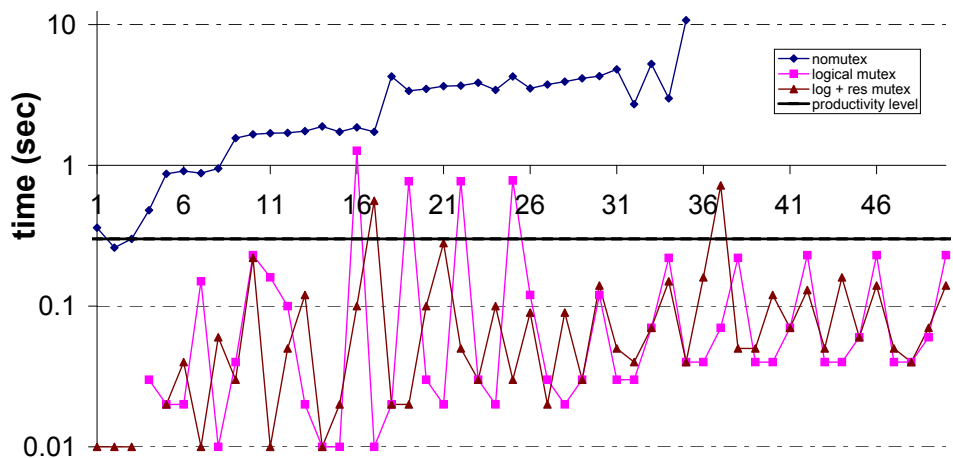

Figure 15: Performance for the current prototype built by us and shown in Figure 1. Note that time is plotted logarithmically.

Figure 14 and 15 show the performance of our planner in two of the most complex parallel printer prototypes built by Xerox and by us. Their productivity levels are higher than any other printer of the same class currently on the market. In each figure, we show the CPU time consumed per-sheet for the basic test of using the planner for a print job of 50 sheets: (1) without mutexes; (2) with serial temporal planning graph with mutexes; (3) combination of logical and resource mutexes; and (4) the baseline requirement for the planner's performance to match with the printer's full productivity. The other prototypes investigated using our planner are either simpler or more complicated but used for theoretical investigations only. For the rest of this section, we will refer to the first printer (results shown in Figure 14) as *Configuration 1* and the other (results shown in Figure 15) as *Configuration 2*.

The *Configuration 1* printer in Figure 14 is the simpler one, with 25 main components (including four print engines), 35 action schemata, and a nominal productivity of 170 sheets per minute, leading to a timing requirement on the planner of around $60/170 = 0.353$ sec for planning each sheet. The shortest possible plan for a simplest simplex sheet contains 8 actions. There are normally more than 10 sheets in flight at a time and thus the planner needs to handle the interaction of around 100 actions. However, the planner is typically intended to plan many sheets ahead so the number of interactions is often much higher. This printer is built and run by our industrial partner. In the figure, we show that without mutex, the planner starts taking more than the base time of 0.353 second around sheet 20 and consistently takes higher than the limit around sheet 40. However, with logical and resource mutexes, the planner consistently returns plans within a much shorter time than required. On average, our planner with logical mutex takes an average of 0.0732 second to find a plan and the combination of logical and resource mutex helps reduce the average planning time to 0.0458 second (1.6x improvement). Without mutexes, the planner takes an average of 0.4191 seconds and can not keep up with the full productivity of the printer.

The *Configuration 2* printer tested in Figure 15 is the more complicated one. There are 212 action schemata and the shortest possible plan contains 16 actions. The printer generally handles more than 20 sheets at a time so the planner needs to regularly reason about the interactions between more than 300-400 actions. The productivity level of this printer is 220 pages-per-minute, which leads to the base running time for the planner of $60/220 = 0.27$ second for planning a single sheet. Because of the wider gap in performance between different versions of the planners, we show timing results for this printer in log scale. Without using mutexes, the planner quickly overruns the time limit after a few sheets and grows to more than 10 seconds around sheet 35, when we stopped the experiment. With mutexes (logical, resource) the planner generally takes less than 0.3 second to find a plan. However, occasionally the planner takes longer. But because it usually plans ahead around 10 sheets before releasing plans to the lower level controller, occasional jumps in planning time don't prevent the planner from achieving the full productivity of the printer in practice. The planner averages 0.1336 second with only logical mutexes and 0.0928 second (1.44x improvement) if used in conjunction with resource mutexes.

The results in Figure 14 and 15 indicate that the average planning time for individual sheets increases with the number of previous sheets. This is due to the fact that the planner generally plans faster than the speed at which the printer can print. Thus, as the number of print requests received increases, the number of plans in the *unsent* queue (i.e., planned for, but not sent to the machine yet) increases. This increases the resource contention and branching factor when searching for a new plan, which leads to the increment in planning time. Eventually, the number of lookahead sheets reaches a point where the planning time equals the planner's productivity and a dynamic equilibrium is reached. The planning time does not strictly increase linearly in accordance with the number of sheets planned, but rather shows an oscillating pattern. This is due to the complex interaction between the on-line processes of planning, freezing time points in the found plan, and plan execution. This can lead to easier planning problems when there are more sheets, depending on how those sheets interact with the sheet that is currently being planned.

While it was noted by Smith and Weld (1999) and other work based on building the planning graph that mutex propagation is costly, this was not our experience. In fact, when the printer is rather empty, the total planning time, which subsumes the graph with mutex building time is less than 0.01 second. We believe that this is due to a simpler mutex propagation rule in our planner and the fact that the sequential plan of each sheet makes all actions mutex at each step. Our resource mutex reasoning time is not as optimized as the logical mutex implementation and can be improved, but it does not seem to be a significant impediment in our intended application.

While the results we have presented indicate that our 'optimal-per-sheet' strategy seems efficient enough, further work is needed to assess the drop in quality that would be experienced by a more greedy strategy, such as always placing the current sheet's resource allocations after those of any previous sheet. Similarly, during a lull in sheet submissions, it might be beneficial to plan multiple sheets together, backtracking through the possible plans of the first in order to find an overall faster plan for the pair together. Sheets that have been planned but whose plans have not been released to the printer represent opportunities for reconsideration in light of the newer sheets submitted more recently.

| # | LPG | | SGPlan | | Hybrid | |
|---|------|--------|------|--------|------|--------|
| | Span | Time | Span | Time | Span | Time |
| 1 | 9.3 | 0.01 | 8.3 | 0.45 | 8.3 | < 0.001 |
| 2 | 13.3 | 0.02 | 9.3 | 308.46 | 9.4 | < 0.001 |
| 3 | 26.6 | 0.08 | - | - | 9.9 | 0.02 |
| 4 | 15.2 | 0.07 | - | - | 10.6 | 0.02 |
| 5 | 21.3 | 0.12 | - | - | 11.1 | 0.03 |
| 6 | 22.4 | 0.23 | - | - | 11.8 | 0.03 |
| 7 | 30.3 | 8.73 | - | - | 12.3 | 0.04 |
| 8 | 19.6 | 52.55 | - | - | 13.0 | 0.06 |
| 9 | 24.2 | 16.69 | - | - | 13.5 | 0.07 |
| 10 | 23.0 | 20.02 | - | - | 14.2 | 0.07 |
| 11 | 29.7 | 40.14 | - | - | 14.7 | 0.08 |
| 12 | 18.3 | 138.53 | - | - | 15.4 | 0.09 |
| 13 | 42.6 | 29.09 | - | - | 15.9 | 0.18 |
| 14 | 34.9 | 427.41 | - | - | 16.6 | 0.21 |
| 15 | 35.3 | 18.95 | - | - | 17.1 | 0.28 |

Table 1: Comparison of LPG, SGPlan, and our hybrid planner, showing the makespan of the plans found ('Span') and planning times ('Time') in seconds for problems with various numbers of sheets ('#').

### 4.4.1 Scaling Against Generic Planners

Although our planner has certain features, such as controllable action durations, that are beyond the capabilities of existing planners, it is still interesting to compare against off-line systems to validate our new approach. If existing generic systems could solve basic printing control problems well, it might be possible to extend them, rather than developing the more specialized planner architecture described above. Therefore, we built a tool to automatically convert our custom domain language into the PDDL2.1 temporal planning language, allowing us to test current state-of-the-art planners.

While our domain must be simplified to fit the limitations of PDDL, we observe that even these simplified problems are not easy to solve by state-of-the-art academic planners such as SGPlan (Chen, Hsu, & Wah, 2006) and LPG (Gerevini, Saetti, & Serina, 2003), winners of the 2004 and 2006 International Planning Competitions. Since both planners cannot solve any problem for the *Configuration 2* machine from Figure 2, we tested them on the much simpler *Configuration 1* machine. While we only tested a monochrome job with up to 15 simplex sheets, this already stretched the limits of LPG and SGPlan. Our planner can plan ahead hundreds of sheets for this machine. As can be seen in Table 1, SGPlan took more than 5 minutes to find a two-sheet plan that only took our planner less than 0.001 second to find. Compared to SGPlan, LPG is much faster, although the quality of the plan LPG finds is much worse. On average, LPG returns plans with 86% longer makespan and is about 400 times slower than our planner. For the objective function of minimizing wall-clock finishing time (which combines planning time and plan makespan), our planner is more than 1000x better than both planners for this small printer configuration.

In addition to being faster, our hybrid planner is also more predictable. LPG's planning time has much higher variance and it sometimes takes longer to plan for a smaller job than a bigger one. For example, it took LPG 22 times longer to plan for the 14-sheet job in Table 1 than it did for the 15-sheet job. This makes it unsuitable for real-time on-line planning, which depends on accurate estimation of planning times for efficient temporal event management.

### 4.4.2 The 2008 International Planning Competition

A version of our printing domain was used in the $6^{th}$ International Planning Competition (IPC6), held in 2008 and the results were presented at the ICAPS-08 conference. This allowed us to evaluate our planner against many state-of-the-art systems. The deterministic part of the competition had three tracks:

1. *sequential* with objective function of minimizing total cost of actions in the plan

2. *temporal* with objective function of minimizing the plan makespan.

3. *net benefit* with objective function of minimizing the trade-off between total goal utility and action cost.

For all three tracks, the emphasis was on finding good solution quality. Thus, planner running time is not part of the overall scoring metric. Specifically, each planner was given 30 minutes to run on a particular planning instance. The cost of the plan returned within the time limit is used to calculate the score for that particular planner in that particular instance. The score for a given instance is *cost of best known solution / cost of generated plan*, where the cost of generating no plan is infinite. There are a total of 30 instances in each domain and thus the maximum score any competitor can achieve is 30 (if all solutions returned have the best quality among all competitors, or equal to the best known solution generated by a specialized solver).

Real-world planners are often demonstrated on complex domains such as spacecraft or mobile robot control which can be difficult to simulate and thus make awkward benchmarks. Most popular temporal planning benchmark domains are off-line in the sense that the planner's speed does not affect solution quality. There remains a need for a simple yet realistic benchmark domain that combines elements of planning and scheduling, especially in an on-line setting. As a step toward bridging this gap, the organizers of IPC6 elected to use the PARC printer domain in two tracks: *sequential* and *temporal*. The temporal track was the most natural fit due to the default objective function of maximizing the printer's productivity, which equals to minimizing the makespan of the plan finishing all print-job requests. For the sequential track, minimizing the total printing cost was used. Each action has a certain cost value and using a more expensive color print engine to print a black&white page costs more than using a monochrome print engine. However, the cost trade-off may not be clear-cut if the feeder, where the blank sheets originally reside, is closer to the color print engine than to the monochrome engine.

Even though the internal representation of our planner, which was used as a starting point for the competition domain description, is not too far from the PDDL representation, there were some difficulties in creating the competition domain file and the problem set.

| $C^3$ | DAE1 | DAE2 | DTGPlan | $FF_a^h$ | $FF_{sa}^h$ | LAMA | PlanA | $SGP6$ | Upwards | baseline |
|-------|------|------|---------|----------|-------------|------|-------|--------|---------|----------|
| 18.00 | 14.45 | 5.80 | 16.44 | 16.00 | 23.00 | 20.93 | 0.00 | 24.39 | 26.91 | 26.53 |

Table 2: Scores from the IPC6 Sequential Satisficing Track

Because of the on-line continual nature of the domain and the fact that constraints such as multiple resource allocations by each action and the sequential finishing order of sheets in the same job caused a blowup in the problem size when using pure PDDL, the organizers had to: 1) remove or approximate certain constraints in the original domain; and 2) model less complex machines. Overall, three different printers were modeled. The first one is the simpler four-engine (two color and two mono) *Configuration 1* machine that we described in Section 4.4. The second one is a stripped down version by using only half (one color and one mono) of the first printer. The third one is another variation with only one mono and one color printer. The first two printers have a rather symmetric design and a third one is asymmetric. All three are significantly simpler than the fixture built at PARC. We helped the IPC organizers model the actions as accurately as possible, and thus even though the printer configurations are hypothetical, they reflect the characteristics of real hardware.

For the problem files, to reduce complexity, only print requests of a single job with multiple sheets were used. The sheets are randomly set to be either simplex (one-sided print) or duplex (two-sided print) and each image is also randomly selected to be either mono or color. The number of sheets varies from 1 to 20. Given this print-job request and a particular printer configuration, the competing planner needed to find a plan with lowest total printing cost in the sequential track (matching effectively between image requirement and print engine capabilities) and smallest makespan in the temporal track (synchronizing effectively between different print engines). In the actual competition, only problems ranging from 1-10 sheets were used for each of the three modeled printers and only simplex sheets were used for the biggest printer (4-engine version) to make the problems not too difficult for a majority of participants.

For all problems used in the two tracks (more details below), we used the planner described in this paper to provide the *best known solutions* to score competing planners. For the temporal planning track, we ran our planner with the default objective function of maximizing the machine's productivity. For the sequential track, we ran our planner with the objective function of minimizing printing cost, as described later in Section 6.1. Given that the plan representation is different between our planner and the standard format used in the IPC, a post-processing step was needed to convert from one format to another. Note that in the plans returned by our planner, there are temporal *buffers* between related time points, such as inter-sheet gaps. The post-processing step does not remove those small temporal buffers, which are not needed for PDDL plans to be valid. Therefore, competing temporal planners could theoretically return shorter makespans than our planner. However, the results as described below show that our planner is still superior to all competing planners in terms of plan quality. The organizers did not officially reveal the plan running time but unofficial results showed that our planner was also much faster than all competing planners in solving most problems.

Tables 2, 3, and 4 show the IPC results in the three sub-tracks in which the PARC printer domain was used: sequential satisficing, sequential optimal, and temporal satisficing. Our

| CFDP | CO-Plan | CPT3 | Gamer | $HSP_0^*$ | $HSP_F^*$ | MIPS-XXL | PlanA | Upwards | baseline |
|------|---------|------|-------|-----------|-----------|----------|-------|---------|----------|
| 3 | 5 | 17 | 0 | 14 | 16 | 7 | 0 | 0 | 10 |

Table 3: Scores from the IPC6 Sequential Optimal Track

| CPT3 | DAE1 | DAE2 | SGPlan 6 | TFD | TLP-GP | baseline |
|------|------|------|----------|-----|--------|----------|
| 17.38 | 11.93 | 6.81 | 11.04 | 5.67 | 1.73 | 13.87 |

Table 4: Scores from the IPC6 Temporal Satisficing Track

planner's score would be 30 for all tracks given that it was used to provide the "best known solutions" for all tracks. The baseline planner for the sequential optimizing track was based on blind search for optimal cost, in the sequential satisficing track was the FF planner, and in temporal satisficing was the Metric-FF planner followed by a greedy scheduler. While the sequential satisficing planners performed well, mostly due to the fact that most problems in the sequential tracks are easy to solve, the competitors did not perform well in the other tracks. The reason that sequential optimal planners did not perform well because they could not solve many problems among those 30 selected. For the temporal satisficing tracks, most planners could solve a large number of instances, but the quality of the plans returned by those plans was not high, thus leading to the low overall scores. In short, the results of the 2008 International Planning Competition reinforced our early study indicating that generic off-line planners are not competitive with our on-line hybrid system in this application. Together, they provide evidence that the demands of our setting warrant a more specialized approach.

## 5. Exception Handling

While maintaining high productivity, and thus high return on investment, is the most common and important objective, it is by no means the only thing that equipment owners care about. To reduce the need for operator oversight and expertise and to allow the use of very complex mechanisms, the system must be as autonomic as possible. Because operators can make mistakes and even highly-engineered system modules can fail, the system must cope with execution failure. This is a crucial part of the RMP value proposition. For example, imagine a printer or copier that never seems to jam, but just runs a little slower as the month goes on. Once a month, someone opens the covers, removes some jammed sheets, and the system is back at full productivity. The RMP systems that our planner is used to control are designed to fulfill this vision of partial productivity when a subset of the modules are down. To make this transition transparent to the user (and thus increase the perceived reliability of the system), we have been concentrating on developing exception handling techniques that minimize user interventions without stopping or slowing down the machine. Current products perform exception handling using rules hard-coded into each machine module. This technique works well for simple straight-line systems, but would be limited to a small predefined subset of failures in more complex topologies. In our modular RMP systems, there are an astronomical number of different printer configurations and failure possibilities, so we require a more general exception handling approach.

In addition, because the system must work with legacy modules in order to be commercially viable, its architecture must tolerate components that are out of its direct control and will give rise to unexpected events. We handle several different exception types such as plan rejection (by the machine controller), model updates (i.e., module's capabilities go on or off-line), and sheet jams.

Since most plans in our system tightly interact through various scheduling and temporal constraints, whether or not they belong to the same print job, an exception affecting any single plan can affect the executability of other plans and the final job integrity. Plans in different stages of their life cycle need to be analyzed and treated differently (see Figure 16). Simple exceptions such as plan rejection and model updates can be handled by discarding recently made plans and rolling back the state of the planner to before those sheets were planned. Our implementation uses non-destructive data structures to make this efficient. Execution failures such as sheet jams require more elaborate handling. While *unsent* plans can be canceled, we need new plans for the sheets that are already in-flight at the time an exception occurs. While this replanning can reuse much of the nominal planning system, it requires some special modifications that we discuss in detail below. In this section, we first provide an overview of the various types of exceptions that we handle and how the plan manager reacts to them; we then concentrate on the hardest part of the exception handling framework: finding a new set of consistent plans for in-flight sheets.

## 5.1 Related Work

There are several previously-proposed frameworks for handling exceptions and uncertainty in plan execution. Markov decision processes (Boutilier, Dean, & Hanks, 1999) and contingency planning (Pryor & Collins, 1996) build plans and policies robust to uncertain environment. Planners built on those techniques are normally slow, especially in a real-time dynamic environment with complex temporal constraints like ours. They are not suitable for our domain where exceptions do not happen frequently, but need to be responded to very quickly. Fox, Gerevini, Long, and Serina (2006) discuss the trade-off between replanning and plan-repair strategies for handling execution failure. Their algorithms work off-line, instead of in an on-line real-time environment such as ours, and they target a different objective function (in their case, plan stability). CASPER system at JPL (Chien, Knight, Stechert, Sherwood, & Rabideau, 1999) uses iterative repairs to continuously modify and update plans to adjust to the dynamic environment. Unlike our system, CASPER uses domain control-rules and thus is less flexible and the replanning decision is also not needed as quickly as in our domain (in our case, sub-second).

## 5.2 Basic Exception Handling

Our planner can handle several types of exceptions. Figure 16 extends the system architecture diagram from Figure 3 and shows in solid lines the possible steps of the replanning process. In general, when an exception occurs, the machine controller sends the planner a message in real time detailing the exception. The planner then cancels plans that have been created but have not been sent to the printer controller to execute. The corresponding goals are rolled back into the unplanned queue. The planner at the same time also tries to find the new plans for sheets that are moving in the printer to avoid further exceptions.

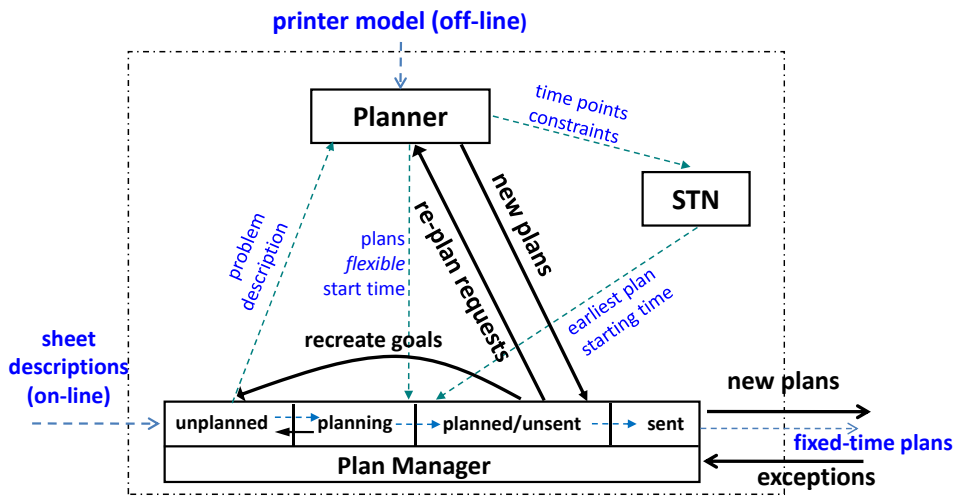

Figure 16: System architecture, showing the steps involved in nominal planning (dashed lines) and replanning (solid lines).

The new plans when found are sent to the machine controller to replace the ones that are executing.

Next, we discuss in detail each of the different exception types.

**Plan Rejection:** When a plan is sent to the machine controller to execute, the controller may reject the plan if one of the relevant modules cannot commit to executing its requested action at the time defined by the planner. While such rejections are rare, they can be caused by module constraints that are outside the scope of the planner's model. For example, a printer engine may need time to bring the toner to the proper temperature—a state variable and constraint not currently modeled in our system. When a plan is rejected, the planner will cancel all plans in the *unsent* queue, in addition to the recently sent and rejected plan. All goals corresponding to those plans will be rolled back to the *unplanned* queue. Even plans that are not directly affected by the error message also need to be canceled and rolled back because those plans were made after the commitments had been made for the rejected plan.

**Module Update:** Machine modules can go *off*-line due to a hardware failure, such as a sheet jam, a benign event, such as running out of paper in a feed tray, or an unmodeled process, such as print engine self-adjustment. Similarly, they can come *on*-line when they are repaired, adjusted, or otherwise made ready. When this happens, the module controller will send a message to the planner indicating which of the module's capabilities is now on/off. If a given capability is turned *off*, then the planner will remove the corresponding action from consideration in future planning episodes. If a given capability is turned *on*, then the planner will add it to the action set for future planning episodes.

**Break-in-Future:** When a module changes the status of some of its capabilities from *on* to *off*, currently executing or unsent plans using that module may become invalid. In this case, the module controller will send messages to the planner indicating which plans are affected. The planner will cancel the affected unsent plans and subsequent plans and move the goals back to the *unplanned* queue. For plans that are executing and thus correspond to sheets that have already been fed into the machine, the planner needs to find new plans for the affected sheets so that they can get to the correct finisher tray without going through the affected modules. The next section describes in detail how to reroute those in-flight sheets.

**Broken:** This type of exception happens when one or more sheets are jammed in the system. The *broken* messages sent to the planner include the ids of all sheets that are jammed and thus cannot be reused or rerouted because of the failure. When some sheets jam, they normally also disable some modules and thus a *broken* message is normally accompanied by several *module update* messages, which are described above. The handling of the *broken* exception is similar to the handling of the *break-in-future* exception in many respects: it involves canceling of unsent plans and finding new plans for the in-flight sheets. However, the main differences are: (1) in-flight sheets that were jammed cannot be rerouted; and (2) more critically, the jammed sheets break print job integrity. We discuss this in detail next.

## 5.3 In-flight Sheet Replanning

In this section, we discuss the problem of finding a new set of plans for in-flight sheets when a sheet is jammed or a module to be used by some plans is broken. The constraints that make replanning more challenging than nominal planning are:

- Sheets cannot stop or slow down while the planner searches for new plans for all in-flight sheets. Thus, if the planner takes too much time to find new plans, the jams and/or module failures will cascade.

- All newly found plans do not have flexible starting times as in the nominal planning case, but should all start from the location where the sheets are projected to be when the plans are found. The new locations depend on the actual replanning time of the planner.

- Any in-flight sheets occurring later in the same print job as a jammed sheet should be rerouted to a *purge* tray. The sheets from jobs without jammed sheets still need to finish in the correct finisher tray and in order.

Replanning involves four main steps: (1) create new goals for the in-flight sheets; (2) predict (an upper bound on) the replanning time; (3) project the sheets according to the original trajectory and the predicted planning time to find their future locations, which will form the new initial state of the replanning problem; (4) find plans for all sheets that are salvageable (those for which it is possible to avoid broken modules and jammed sheets in time), satisfying the constraints listed above.

### 5.3.1 Example

Here we provide a concrete example illustrating our replanning procedure. Figures 17 & 18 show a scenario in which there are three in-flight sheets: $S_{1.1}$ and $S_{1.2}$ belong to the same print job and were planned to go to finisher 2 (in the middle); sheet $S_{2.1}$ belongs to a different print job and is scheduled to go to finisher 1. The third finisher is the purge tray. The original routes are indicated by the dashed lines in Figure 17. Assume that $S_{1.1}$ is jammed. According to the original routes, we have: (1) $S_{1.2}$ will arrive in the finisher tray out-of-order (because $S_{1.1}$ did not arrive before it); (2) $S_{2.1}$ will crash into the module where $S_{1.1}$ jammed. Therefore, we need to find new plans for those two sheets so that $S_{1.2}$ will instead go to the purge tray and $S_{2.1}$ goes around $S_{1.1}$. Finding those plans takes time and given that we cannot stop or slow down $S_{1.2}$ and $S_{1.2}$ while finding the new plans for them, those two sheets will continue their original trajectories to the new locations, which are circled in Figure 18. From there, the machine controller will apply the new plans, indicated in the figure by solid lines, in order to guarantee print-job integrity while avoiding further cascading failures. After the replanning is done, the planner will generate fresh plans to re-create $S_{1.1}$ and $S_{1.2}$.

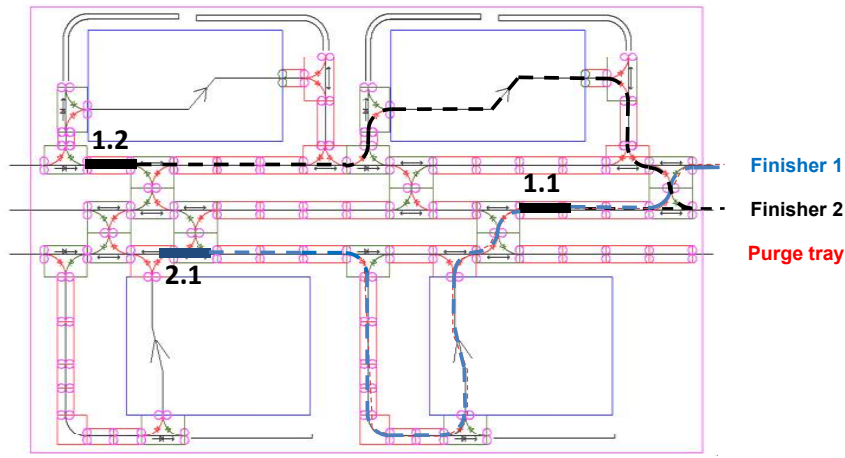

Figure 17: Replanning Example (before jam): sheet 1.1 and 1.2 are planned to enter finisher 2, and sheet 2.1 to finisher 1.

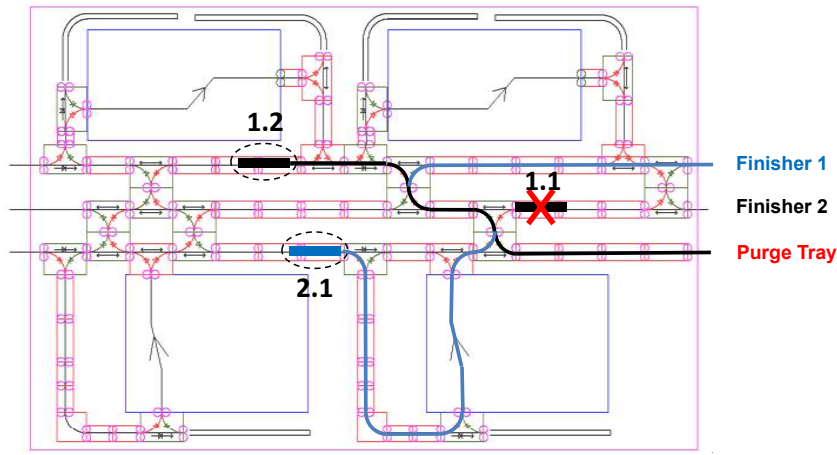

Figure 18: Replanning Example (after jam): sheet 1.1 is jammed, which requires the planner to reroute sheet 1.2 to the purge tray and reroute sheet 2.1 to circumvent the jammed sheet before going to finisher 1.

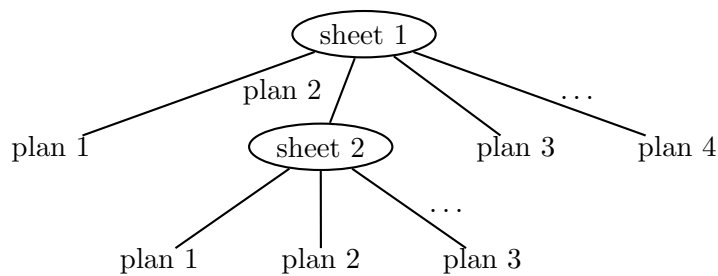

Figure 19: Chaining many searches together gives a search tree with potentially infinite branching factor.

The example above shows one replanning strategy where the new goal for out-of-order sheet $S_{1.2}$ is set to go to the purge tray. This is the default strategy in our replanner that tries to clear out the machine and finish the replanning process as quickly as possible to return to normal operation. However, there are scenarios where the printing media is expensive or the content is confidential and purging sheets is not desirable. In those scenarios, we have also experimented with a different strategy that does not purge $S_{1.2}$ but keeps it in the machine (for example, by looping it in a holding pattern) while waiting for $S_{1.1}$ to be reprinted, then $S_{1.2}$ is routed to the original finisher. The only modification necessary to implement this strategy in our system is to change the way replanning goals and end-time constraints are generated. We have tested this strategy successfully for a small number of sheets, although more sheets could be saved if one were allowed to slow down the transports.

### 5.3.2 Chained BFS

For normal operation, the planner uses A* to find the plan for a given sheet that can end soonest, given the (temporally flexible) plans for the previous sheets. A plan always exists if scheduled sufficiently far in the future. For rerouting, the problem is different. We must find jointly feasible plans for as many in-flight sheets as possible. We cannot greedily plan one sheet at a time, committing irrevocably to the plans for all previous sheets, because the plan selected for one sheet might render subsequent sheets infeasible. This cannot happen during nominal planning, as later sheets are always feasible when scheduled sufficiently far in the future. When replanning, however, we are forced to confront a true multi-body planning problem.

We considered two strategies to solve this problem. The first was to simply plan in the joint action space of all sheets. This would result in a large branching factor and it was not clear to us how to design an effective heuristic evaluation function. We chose a different approach, in which we can retain the view of planning for each sheet individually using heuristic search. However, we overlay an additional search on top of this, as depicted in Figure 19. In the high-level search, a branching node represents the situation in which we have selected certain specific plans for all previous sheets and it is time to select a plan for an

**ChainedBFS** (*problems*)
1. if *problems* is empty, return success
2. $p \leftarrow$ remove first problem from *problems*
3. initialize *openlist* for $p$
4. repeat until *openlist* is empty or node limit is reached:
5.     $n \leftarrow$ best node on *openlist*
6.     if $n$ is a goal, call ChainedBFS with remaining *problems*
7.     expand $n$, adding any children to *openlist*

Figure 20: Sketch of Chained Best-First Search with a depth-first strategy.

additional sheet. The children of that node represent commitments to the different possible plans for that additional sheet. By considering different paths in the high-level search tree, we can consider different combinations of plans for the different sheets. We call this approach *chained best-first search*. In our current implementation, sheets are replanned in their original order, as an approximation of increasing 'distance from exit,' which correlates with increasing flexibility. An alternative approach is to replan in the order of 'urgency' defined as the time left to reroute a sheet before it becomes unsalvageable.

Because the children of a node represent the possible plans returned by a best-first search, the children are not available all at once. Instead, an individual sheet-level planning search will encounter goal nodes one at a time. We cannot terminate the search when we find the first goal node for single-sheet planning because we have no guarantee that the sheet-plan reaching that first goal will make the subsequent sheets feasible. Finding a plan for a single sheet merely results in a new branch in the high-level space, and to retain completeness we must retain the ability to continue our search and uncover additional possible plans. In fact, in printers such as ours that contain loops in the paper path, there may be an infinite number of possible plans for a given sheet. Fundamentally, the high-level search must explore a tree where nodes are expanded incrementally and the branching factor is potentially infinite.

We identified three possible strategies for searching a tree with infinite branching factor. The first is a best-first approach, in which one formulates a traditional heuristic evaluation function for the high-level nodes. These nodes represent commitments to complete plans for a subset of the in-flight sheets, so the heuristic function needs to estimate the probability that those plans will allow feasible plans for the remaining sheets to be found. The infinite branching factor could be handled using Partial-Expansion A* (Yoshizumi, Miura, & Ishida, 2000), although this would require a non-trivial lower bound on the heuristic value of the plans that have not yet been found. It was not clear to us how this might be done. The second possible strategy we considered was limited discrepancy search (Korf, 1996). Unlike depth-first search, limited discrepancy search doesn't necessarily visit all the children of a node, which are potentially infinite for us. The disadvantage to this method is that, because we revisit each node many times with different discrepancy bounds, we will suffer considerable node regeneration overhead.

The third strategy, and the one we used in our implementation, is perhaps the simplest: depth-first search. Figure 20 shows a pseudo-code sketch. Because we have a fixed number of sheets to replan, the high-level search tree has bounded depth. To cope with the potentially infinite branching factor, we impose a limit on the number of nodes each low-level sheet planning search may expand. This avoids the danger of searching forever at one high-level node without finding another goal, and is reminiscent of iterative broadening (Ginsberg & Harvey, 1992). To guide the sheet-level planning, we use a heuristic that minimizes plan duration. This attempts to minimize resource use in the machine and maximize the probability that other sheets will have feasible plans.

## 5.4 Evaluation

Until now, the exception handling strategies in current production printers have been to: (1) stop the production and ask the operator to remove all sheets or (2) use machine-specific customized local rules to purge sheets in the system. Our work is the first to demonstrate automatic exception handling that does not rely on machine-specific control rules.

The planner can handle the two easiest types of exception: *Plan Reject* and *Module Update* without any difficulties. For the *Break-In-Future* and *Broken* exceptions, we can currently reroute on the fly up to five sheets for the machine shown in Figure 1. This number may seem low, but recall that replanning is harder than nominal planning by a factor exponential in the number of in-flight sheets. For the simpler prototype systems at Xerox with fewer (but larger) modules, four print engines, and an aggregate throughput of 180 pages-per-minute, our planner has been able to successfully reroute all reroutable sheets when different jams happen. We have demonstrated our replanning technology in real-time by allowing people come up and either turn on/off modules, or jam sheets intentionally, sometimes right before sheets hit the broken module. Upon receiving the error messages from the machine controller, the planner is fast enough to reroute the sheets around the failed modules or jammed sheets to the correct locations. In addition to experimenting with the physical hardware built at PARC and by Xerox, we have also tested replanning in simulation, by connecting the planner to a visualizer instead of the machine controller. The on-line appendix contains two videos of the planner performing in-flight rerouting on the PARC prototype, both in simulation (Video 3, 'replanning simulation') and in hardware (Video 4, 'replanning hardware').

In addition to testing our replanning framework on different hypothetical printer configurations and different fault modes, we have also investigated different exception handling strategies. For example, when the printing media is expensive and the replanning objective function is switched from the default objective function of finish replanning as quickly as possible (which can lead to many purged sheets) to saving as many sheets at possible (which can lead to longer replanning time) then the planner has been able to successfully route up to 2 out-of-order sheets in long routes (that may contain loops) in the system waiting for the jammed sheet to be printed before being routed to the correct finisher tray. While achievement of replanning for up to five sheets in a large RMP machine may not seem very impressive, we want to point out that: (1) our planner can reroute all reroutable sheets in simpler machines (which is still much more complex than the biggest multi-engine printer Xerox currently has on the market); (2) the large machine is very complex for automated planning—the last two IPC winners SGPLan and LPG cannot even find plan for a single sheet in nominal planning using the PDDL2.1 version of our printer domain.

## 6. Handling Multiple Objectives

Our second major extension to nominal planning is aimed at better meeting shop owner's needs in the nominal case. Up to this point, the planner's objective has been to run multi-engine reconfigurable printers at full productivity, optimizing for machine throughput. Productivity, while very important, is only one of the many optimization criteria that naturally exist in real-world planning and scheduling applications like the printer control domain. In this section, we will describe several additional objective functions that were pointed out as important by our industrial partner, and discuss how we extended our planning framework to handle them.

In a modular system with multiple print engines, one might want to optimize the cost of printing by choosing to print black-only pages only on monochrome engines and avoid using more expensive color engines. Also, one might want to optimize image quality by choosing to print pages from the same document only on print engines whose current marking gamuts are similar. The printer controller needs to give operators the ability to trade off these conflicting objectives while maintaining robust operation. We meet these challenges using (1) an optimization objective that combines separate estimates of productivity and printing cost, and (2) multiple heuristic look-ups to efficiently handle image quality consistency constraints. In contrast to an explicit multi-objective optimization, in which a planner would return an selection of non-dominated solutions on the Pareto frontier, presumably for a human to choose from, our planner needs to select a single solution for execution, so we need to combine the multiple criteria into single objective. Because our planner is built atop generic state-space heuristic search, we need only design a new comparison function to order search nodes. In addition to linear combinations of objectives, it is relatively easy for us to handle tiered criteria using tie-breaking strategies.

There are several academic domain-independent planners such as GRT (Refanidis & Vlahavas, 2003) and LPG (Gerevini, Saetti, & Serina, 2008) that can optimize for multiple objectives or trade-off between planning time and plan quality. Standard planning languages, especially PDDL3 (Gerevini & Long, 2006), allow specifying complex objective functions in the weighted-sum format (as in our framework). While our planner is also based on domain-independent planning technology and uses an extension of PDDL, it works in a dynamic on-line continual environment and interacts with a physical machine, not in an off-line abstracted environment like previous planners.

### 6.1 Optimizing for Printing Cost

For systems with heterogeneous print engines, the cost of printing a given page depends on which of the engines is used. For example, it is generally costlier to print a black-and-white page on a color engine than a monochrome one. Thus, to minimize the overall printing cost, one should use the engines with the lowest printing cost that still satisfy the image type and quality requirements of a given print job. By doing so, only a subset of all the available engines will be used for printing a job and thus the overall productivity may be reduced.

To strike a balance between machine productivity and printing cost, we have implemented an objective that can trade off productivity for cost and vice versa. We show that by combining different performance criteria into a single objective, the same optimization framework that works so well for single-objective planning can be efficiently applied to the multi-objective case. Below are the main steps required to extend the planner from supporting a single objective to multiple objectives.

**Step 1:** Extend the planner's representation of machine capabilities to model action cost. Specifically, we added a cost field representing the cost of executing each capability. In addition, there is an overall objective field with user-supplied weights for each of the two objectives: $obj = \min \ w_1 * t + w_2 * c$, where $t$ is the end time and $c$ is the accumulated total cost of printing all sheets.

**Step 2:** Create one heuristic estimation function for each of the objectives. To find the best route for a given sheet, we estimate how good a potential route is according to each of the objective functions. Finishing time is estimated using a temporal planning graph adjusted with resource conflicts, as described in Section 4.3.4. To estimate the total plan execution cost, we use dynamic programming starting from the initial state (i.e., sheet in the feeder) to compute the total cost to reach different reachable states. The computation is similar to cost propagation on the planning graph as in the Sapa planner (Do & Kambhampati, 2002).

**Step 3:** Extend the search algorithm to considering multiple objectives simultaneously. The estimations on total time and cost are combined using the user-supplied weights (as described in Step 1) to compare nodes in the best-first A* search algorithm. Given that both heuristics for time and cost are admissible, like the single objective planner, our planner is guaranteed to find an optimal solution for any given sheet. If the weights are not given, the planner chooses to prioritize the objectives. For example, the planner can first find the plan that has the lowest cost, and then break ties favoring plans with higher productivity, then favoring one with lower wear and tear, and so on. This mechanism has been implemented and fully integrated into our planner. The default option when no weights are specified is to optimize for productivity and break ties on total cost.

## 6.2 Planning for Image Quality Consistency

Maintaining image consistency across a set of heterogeneous print engines is especially important for a multi-engine printing system. The planner achieves this by enforcing additional image-consistency constraints while searching for an optimal plan. In color science, the (in)consistency of two colors is measured by a function, often denoted $\Delta E$, that calculates the distance between the two in some device-independent color space. While there exist a variety of such functions (the most popular of which is called $\Delta E2000$; see Green, 2002), for our planning purpose it suffices to assume that given any two engines, a $\Delta E$ function returns a non-negative real-valued scalar, called $\Delta E$ distance, that measures the discrepancy in *perceived* color as a result of printing the same image on these two engines. Because facing pages (i.e., pages that face each other in a bound book or magazine) are most sensitive to image-consistency issues, we thus consider the following constraints in our planner:

1. *facing-page constraints* that require the facing pages of a job be printed by the same print engine

2. $\Delta E$ *constraints* that allow only engines within some maximum $\Delta E$ distance to print facing pages

Given that in reality no two engines can have a $\Delta E$ distance of zero, the facing-page constraints can be viewed as a special case of the $\Delta E$ constraints with the maximum $\Delta E$ distance set to zero. Thus, we only need to focus on the latter, which is more general. To enforce $\Delta E$ constraints, the planner keeps track of the set of print capabilities that can be used to print the front side of a sheet, which is constrained by the print action applied to the back side of its previous sheet. Since the first sheet of a job does not have a previous sheet, the set of print capabilities eligible for printing its front side is unconstrained (i.e., equal to the entire set of print capabilities). For subsequent sheets of the same job, however, only a subset of print capabilities is allowed. Such a subset is computed based on the $\Delta E$ constraints by including only capabilities of those engines whose $\Delta E$ distance to the print engine that printed the back side of the previous sheet is less than or equal to some maximum distance. In most cases, this has to be determined on-line, because the $\Delta E$ distance between a pair of engines can drift over time. Thus, our planner maintains an on-line version of a pairwise $\Delta E$-distance matrix for all the engines in a printer.

While adding extra image-consistency constraints can reduce the brute-force search space (if the constraints make the set of reachable states smaller), in practice we found this often makes the search problem *harder*, because the heuristic computed for the unconstrained problem, while still admissible, is no longer informative. To improve the accuracy of the heuristic, the planner computes the temporal planning-graph heuristic for all legal combinations of print capabilities that can be used to print one side of a sheet, and then stores them in multiple lookup tables, one for each combination. When a heuristic estimate for a search node is needed, the planner calculates an index into the lookup table based on the state description (e.g., sheet location, monochrome or color printing), in much the same way as lookups are done in pattern databases (Culberson & Schaeffer, 1998). In our implementation, a hash table of hash tables is used to store multiple lookup tables, but for any given sheet only the relevant hash table(s) is loaded before the sheet is being planned, because the set of eligible print actions is known and fixed at that time.

Since there are only a limited number of ways of printing a single face of a sheet, this approach to improving heuristic accuracy has little overhead yet can significantly reduce the time it takes to find an itinerary. Interestingly, the same approach can also be used to improve the accuracy of the heuristic for handling exceptions in which jammed sheets block the paper paths to some engines, because then only unblocked engines are eligible for printing sheets, creating planning problems that are similar to enforcing the $\Delta E$ constraints. For example, one can set the $\Delta E$ distance to any blocked engine as infinity, which effectively forces sheets to go through only unblocked engines, and the computational savings comes from the use of a more accurate heuristic that is built specifically for a particular set of unblocked engines, instead of a nominal-case heuristic that assumes no engine is blocked.

### 6.2.1 PLANNING WITH A CONSTRAINED ACTION SET

From an algorithmic perspective, our approach to planning for image-quality consistency corresponds to solving a constrained planning problem with a reduced set of actions (compared to its unconstrained version). Given a planning problem with $k$ actions, one can create $O(2^k)$ different versions of the constrained problem. Thus, pre-computing the temporal planning-graph heuristic for all possible subsets of actions can quickly become infeasible as $k$ increases. Here we describe a general solution that strikes a balance between heuristic accuracy and the space overhead for storing multiple lookup tables, one for each subset of the actions. The idea is to limit $m$, the maximum number of actions that are removed from the unconstrained problem, and compute heuristic lookup tables only for those constrained problems. For example, it is usually feasible to enumerate those constrained problems in which only one or two actions are removed from the action set. To compute the heuristic value of a state in a constrained problem that is not included in this pre-computed set, the algorithm consults all the lookup tables whose removed actions form a subset of the actions removed in the constrained problem, and returns the maximum value as the heuristic estimate of the state, since the value returned by any of the lookup tables is admissible.

More formally, let $h(s|P)$ be an admissible heuristic estimate for state $s$ in the constrained problem with the set of actions $P \subseteq A$ removed from the original action set $A$, and let $m$ be the maximum number of actions removed in any constrained problems for which the heuristic is pre-computed. The heuristic estimate $h(s|P)$ can be calculated as follows,

$$h(s|P) = \begin{cases} h(s|P) & \text{if } |P| \leq m \\ \max_{Q \subset P \,\wedge\, |Q|=m} \, h(s|Q) & \text{otherwise} \end{cases}$$

The new heuristic resembles the $h^m$ family of admissible heuristics (Haslum & Geffner, 2000), where $m$ limits the maximum cardinality of the set of atoms considered in the construction of the heuristic. The difference is that our heuristic considers the set of removed actions, whereas the $h^m$ heuristic considers the set of satisfied atoms. Our heuristic can also be seen as a kind of multiple pattern database (Holte, Felner, Newton, Meshulam, & Furcy, 2006) in which one can take the maximum over a set of heuristic estimates without losing admissibility, although ours is based on action-space abstraction and (multiple) pattern databases are based on state-space abstraction.

## 6.3 Evaluation

To test the ability of our planner to trade off between machine productivity and printing cost, we have tested on the model of a four-engine prototype printer built at Xerox. This is a better test bed for this trade-off investigation because that printer has a mixed set of printer engines (two color and two black-and-white engines) instead of four identical black engines such as in the PARC prototype system. Moreover, the engines are aligned asymmetrically and thus the paths leading to different engines are slightly different. We have modeled the costs for all different components in consultation with Xerox engineers. We are especially interested in modeling the cost to print black pages on different engines: printing them on more expensive color engines costs more than on cheaper monochrome engines. By varying the weights between the two objective functions, we have been able to show that: (1) increasing the weight given to productivity results in more printer utilization of all four engines; (2) increasing the weight on saving printing cost leads to reductions in the number of unnecessary costly printing, thus fewer black sheets are printed on color engines. We can observe the trade-off between modules with similar functionality as well, such as between different feeders, finishers, or paper paths. For example, increasing the weight for saving costs lowers the number of sheets fed from a more expensive but faster feeder. We have also tested our search on other hypothetical printers with mixed components and similar results were observed. We also observed that moving from single to multiple objectives did not slow down our planner and thus did not affect the overall productivity.

We also tested the performance of our planner on image-consistency planning. The model of the printer used has four monochrome engines, two of which are faster but low-quality engines, and the remaining two are slower but high-quality engines. All four engines are connected through asymmetric paper paths. We ran the simulation with a 20-sheet job that requires using the two high-quality engines for double-sided printing. This can be done with certain $\Delta E$ constraints, which can prevent the planner from choosing the two low-quality engines. Since we are particularly interested in the effect of the heuristic on the search performance, we tested the planner with and without using multiple lookup tables, which made a significant difference in the number of node expansions in A* search and planning times. On average, when the multiple lookup table heuristic is used, the planner expands only 1783 nodes per sheet; whereas using the heuristic computed for the unconstrained problem, which grossly underestimates the remaining makespan for constrained problems, needs 6458 node expansions to find a plan. In terms of running time, the one that uses multiple lookup tables is 60% faster than using the naive heuristic.

One future direction is to investigate a different objective entirely: *wear and tear*. Under this objective, one would like the different machines in the plant to be used the same amount over the long term. However, because machines are often cycled down when idle for a long period and cycling them up introduces wear, one would like recently-used machines to be selected again soon in the short term. Although our implementation currently only supports throughput and cost, it should be easily extensible to support additional objectives.

## 7. Deployment

In the process of building and deploying the planner, we utilized many off-the-shelf techniques from academic research in planning, extending, and integrating them to form a fast on-line planner/scheduler. In this section, we list the most important lessons we learned and describe the ancillary tools that were necessary to develop and deploy the planner. We hope that they can be useful for both application developers and academic researchers in planning.

### 7.1 Lessons Learned

**Modeling is important.** We mean this in two respects. First, it was important to our end-users that printers were modeled with a specialized representation in which machine modules and the connections between them are the main themes of the language. As we discussed in Section 4.1, this representation is then compiled into the planner input language, taking the capabilities of different modules along with their inter-connections and producing action schemata. In this process, the set of machine capabilities are compiled into a higher number of action schemata that ground some of the parameters. Through discussion with our users and industrial partners, we feel that the machine-centric language involving modules, machine instances, and inter-connections is easier for them to understand and accept, while the compiled-down representation makes it much easier for us to adopt STRIPS planning techniques.

Second, we also found that, because we understood the search algorithm (regression with three-value state representation) and the heuristics (planning graph with mutexes) used by the planner, we could manipulate the modeling of the actions, goals, and initial states to produce quite different computational results. Consider a simple example of the action $a_{12} = move(l_1, l_2)$ that moves an object from location $l_1$ to $l_2$. The most common form of STRIPS representation of this action is $Pre(a_{12}) = \{ at(l_1) \}$ and $Effect(a_{12}) = \{\neg at(l_1), at(l_2)\}$. Recall that we use the three-value representation in which a literal can have values *true, false*, or *unknown*. Regressing the (partial) state $s_1 = \{at(l_2)\}$ using $a_{12}$ will get us to state $s_2 = \{at(l_1), Unknown(at(l_2))\}$ which is regressable through both actions $move(l_3, l_1)$ and $move(l_3, l_2)$. While the normal regression rules may not consider $move(l_3, l_2)$ because it will not lead to an optimal length plan, regressing $s_2$ through $move(l_3, l_2)$ will not cause inconsistency. In fact, in a domain like ours, we do branch over all regressable actions because there are scenarios in which we need to buy time to free up resource allocations. Note that in this example, sophisticated techniques to discover invariants such as TIM (Fox & Long, 1998) or DISCOPLAN (Gerevini & Schuert, 1998) can discover that the object can only be at a single location at any time and thus that $s_2$ is not regressable through action $move(l_3, l_2)$. However, we can eliminate that branch simply by adding a precondition $\neg at(l_2)$ to the action description of $a_{12}$ and make sure that in the goal state, any location other than the goal is *false*. Generating an extra child node and propagating all constraints in our domain is expensive because, in addition to a logical part, our state representation includes temporal and resource databases. So cutting down on the number of generated nodes is important, and this can be partially accomplished by careful modeling.

Similarly, adding or removing predicates from the domain description can have a great effects on the heuristic estimation derived by the planning graph through mutex propagation. We experienced a scenario in which adding two extra predicates representing subgoal completion and modeling the domain slightly differently achieved a speedup of nearly 10x for some printer configurations.

These manipulations are reminiscent of the work of Rintanen (2000), who showed how domain advice expressed in linear temporal logic, such as 'don't move a package that is at its destination', could be compiled into the planning operators of a domain using conditional effects, leading to great gains in planning efficiency. However, we want to emphasize that in a highly configurable systems like ours, it can be dangerous to encode explicit action choices or pruning into the domain. It can be hard to guarantee that completeness or optimality will be maintained under all possible job mixes or failure combinations. (For example, looping a sheet may free up a resource that will allow the job to complete earlier.) Our approach is to encode the domain 'physics', that is, things that are universally true in the domain and help keep the search within reachable states, but not any control rules in the sense of heuristic action selections, such as 'when *condition*, choose *action*.' Our point is that the same physics can be represented differently, even if limited to STRIPS, and finding the right match with the chosen search strategy can dramatically affect the planner's performance. As application developers, not having to work with a fixed benchmark domain representation allows us to exploit another dimension in modeling to improve our planner's performance.

**The most suitable planning algorithm depends on the application specifications.** Even after formulating our domain using an extension of STRIPS, we went through several implementations of different planning algorithms before settling on the current one. Our first version was a lifted partial-order planner, which we still think is the more elegant algorithm. We then implemented a grounded forward state space planner, because that approach has dominated the planning competitions. However, as discussed in Section 3.1, we realized that a combination of the constraint that sheets in a same print job should be finished in order and our objective function of minimizing the finishing time is not suitable for forward state-space search. We finally settled on a backward state-space framework, which is much faster in our domain. The lesson we drew from this is that just because some approach works best in a wide range of benchmark domains in the competition does not mean that it is the best choice for a given application; and if it doesn't work, it does not mean that other less popular approaches cannot do significantly better. Therefore, understanding your domain, the important constraints involved, your objective function, and how different planning algorithms work can help in selecting the most suitable strategy. Looking up competition results is not a replacement for understanding the variety of applicable planning algorithms.

**Having a fast and robust temporal reasoner is very important.** In our planner, even though the source code for the Simple Temporal Network (STN) totals less than 200 lines of code, it is critical in handling all temporal relations between actions and resource allocations within a single plan and between different plans. In a real-world application where there are various temporal constraints and delays to take into account, such as communication, setup time, machine controller coordination, and time synchronization delays between the planner and the other components in the overall control architecture, ensuring temporal consistency is one of the most important tasks necessary for keeping the planner running without interruption for a long period of time. Having an explicit temporal reasoner also helped us to uniformly represent and manage start and end times of actions and different types of resource allocations. It also allowed us to smoothly extend from handling fixed duration actions to action with variable durations, and extend from regular resource allocations to resource allocations caused by external events such as cyclic resources allocation by uncontrollable processes. In our domain, variable action durations are context-independent and are different from actions such as *refuel* in the Logistics domain used in the competition. We haven't noticed many planners in the competition having an explicit general-purpose temporal reasoner, except IxTeT(Ghallab & Laruelle, 1994). However, we would like to emphasize that in a real-world setting in which the planner needs to coordinate with other software and expects to face various time constraints and delays, this is critical.

**There are many uses for a planner.** Besides its main job of controlling different printers, the planner has also been used extensively for system analysis purposes. Thus, the planner is tested against (1) different printer designs to help decide the better ones; (2) printers with various broken modules to test the reliability of each printer. Those analyses can help the product group to decide which printer to built for a given purpose. For example, our customer ran an extensive test consisting of 11,760 different planner runs for variations of a single printer configuration. Among those runs, they used the planner to test different combinations of possible broken points, different print-job mixes, and changed speeds of different modules. Another use has been to test the performance of the upstream job submission and sequencing methods. The most direct and accurate way to evaluate a job sequencer is to run a long print-job mix (thousands of sheets or more) through the planner and measure the total makespan. Recently, we completed a print-job mix of 50,000 sheets without any break, which is more intensive than the regular real-life printer operation. Through these experiences, we learned that there are many potential applications of a planner beyond direct machine control.

**Exception handling.** Given that the planner interacts with other parts that are either higher or lower in the control hierarchy, exceptions can come in many forms. We believe that similar exceptions would occur in most applications where the planner interacts with physical world. While robust exception handling (such as replanning) is important, we found that there is much less research on this topic compared to other branches of domain-independent planning.

We hope that these observations can help researchers to develop planning techniques that are closer to those needed in real-world applications and that they are also useful for those considering deploying AI planning in their applications.

## 7.2 Ancillary Tools

In the course of building our system, we developed a number of ancillary tools around our core planning and scheduling software. Among these tools, the most notable piece is the *visualizer*, which simulates the movement of each sheet inside the printer in real time. Like the planner, the visualizer adopts the same model-based principle to make it as machine-independent as possible. Because an itinerary is given as a discrete sequence of actions, each having a single time stamp that prescribes the start time of the action, linear interpolation is used to compute the position of a sheet when the current simulation time is somewhere in between the start times of two consecutive actions. The visualizer works in one of the following two modes: the on-line mode that accepts live itineraries sent by the planner over sockets, and the off-line mode that reads in previously recorded itineraries from a file stored on disk.

To separate the visualization engine from the specific designs of a printer, we developed a simple module definition language for describing the dimensions of each module type, the locations of input and output ports within a module's local coordinate system, the travel distance between a pair of input and output ports, and optionally a customized drawing function that can be used to render the type of modules on the screen. Besides the definition of module types, the visualizer needs to know the location as well as the orientation of each module in a machine-wide coordinate system. While it is possible to specify all these information manually, we developed another ancillary tool called the *visualizer pre-processor* that can be used to automate this laborious yet error-prone task. With this tool, the user only needs to specify the location and orientation for one module, called a *seed module*, from which the locations and orientations of all the other (directly or indirectly connected) modules are deduced based on the connectivity graph of the modules. For machines with more than one feasible configuration, our tool can find all possible solutions and store them in multiple files that can be used later by the visualizer. Besides the nominal case, the visualizer can also simulate various exceptions such as paper jams and break-in-future scenarios. Our long-term vision is for the visualizer to become a design, debug, and verification tool for the manufacturer, as well as a GUI console for the end user who operates the printer.

To make it easier to run tests on our modular printers, we also developed a wrapper program that 'glues' together the planner and the controller (or the visualizer). It takes as input a set of pre-defined test scenarios specified with succinct syntax (e.g., 10sc means print 10 single-sided color sheets). To support the simulation of pre-fabricated exceptions, it sends special messages to the visualizer that contains information about when or where a sheet jam should occur. It also supports simultaneous printing jobs for printers with multiple finishers, and uses a round robin algorithm to draw sheets from jobs at the same rate to maintain fairness. To facilitate remote testing and debugging, the wrapper program uses sockets to communicate with the machine controller (or the visualizer).

## 8. Conclusion

We described a real-world domain that requires a novel on-line integration of planning and scheduling and we formalized it using a temporal extension of STRIPS that falls between partial-order scheduling and temporal PDDL. We presented a hybrid planner that uses state-space regression on a per-sheet basis, while using a temporal constraint network to maintain flexibility through partial orderings representing resource conflicts between plans for different sheets. Our system has successfully controlled three hardware prototypes and outperforms state-of-the-art planners in this domain. No domain-dependent search control heuristics are necessary to control a printer composed of 170 modules in real time. We described extensions to handle two critical issues: (1) real-time execution failures; and (2) objective functions beyond productivity. We have successfully demonstrated our fast replanning and multiple objective handling on three physical prototype printers and many other potential printer configurations in simulation.

Our work provides an example of how AI planning and scheduling can find real-world application not just in exotic domains such as spacecraft or mobile robot control, but also for common down-to-earth problems such as manufacturing process control. The modular printer domain is representative of a wider class of AI applications that require continual on-line decision-making. Through a novel combination of fast continual temporal planning techniques, we have shown how artificial intelligence can successfully enable robust, high-performance, autonomous operation without hand-coded control knowledge.

## Acknowledgments

Much of this work was done while the first author was with the Palo Alto Research Center. Preliminary results from this project were published by Ruml, Do, and Fromherz (2005), Do and Ruml (2006), and Do, Ruml, and Zhou (2008) and summarized by Do, Ruml, and Zhou (2008). The authors would like to thank the members of the Embedded Reasoning Area at PARC, especially Lara Crawford, Haitham Hindi, Johan de Kleer, and Lukas Kuhn, as well as Danny Bobrow, David Biegelsen, Craig Eldershaw, and Dave Duff for their help and contributions to the project. Our industrial collaborators not only provided domain expertise but were invaluable in helping us to simplify and frame the application in a useful way. We'd like to especially thank Bob Lofthus and Ron Root for their enthusiasm and perseverance and Steve Hoover for supporting the project.

## Appendix A: Video

The on-line appendix on the JAIR website contains four movies of the system in action:

1. **nominal-simulation.mp4:** shows one simplex job of 200 sheets being run in a simulation of the PARC prototype printer shown in Figure 1. The planner keeps all four print engines busy, achieving full productivity of the system.

2. **nominal-hardware.wmv:** shows two simplex jobs being run simultaneously using all four engines of the PARC hardware prototype. The two feeders are on the left and two simple finishing trays are on the right. Red lights on the machine modules show the position of sheets. (Background time synchronization is indicated by the periodic blinking.) In the lower left corner, a schematic visualization shows how sheets are moving through the machine, with one job colored blue and the other red.

3. **replanning-simulation.mp4:** show a simple exceptions handling scenario in simulation. Blue sheets and red sheets belong to different jobs. The second sheet of the blue job jams. The third sheet, already in-flight, is rerouted to the middle 'purge' tray and fresh plans are initiated to recreate both sheets. The red job continues uninterrupted.

4. **replanning-hardware.wmv:** demonstrates two exception handling scenarios. The first shows simple on-line replanning. After a sheet has launched, a button is pushed on a module that the sheet is headed toward, to mark the module as 'broken.' This initiates replanning, and the sheet is routed around the 'failed' module. A second module's button is pushed, marking it failed and thereby blocking the finishing tray that the sheet was headed toward. The sheet is rerouting again and emerges at the remaining finishing tray.

   In the second scenario, the module that is 'broken' already contains the first sheet of a two-sheet job. The replanner is fast enough to reroute the second sheet around the jammed first sheet to a purge tray. The original two sheets are then planned again from scratch and arrive at the lower finishing tray.